\documentclass[final]{cvpr}

\usepackage{times}
\usepackage{epsfig}
\usepackage{graphicx}
\usepackage{amsmath}
\usepackage{amssymb}
\usepackage[ruled,vlined]{algorithm2e}
\usepackage{booktabs}
\usepackage{graphicx}
\usepackage{subfigure}
\usepackage{booktabs}
\usepackage{multirow}
\usepackage{makecell}
\usepackage{enumerate}
\usepackage{amsmath,amssymb} 
\usepackage{color}
\usepackage{arydshln}
\usepackage{pifont}
\usepackage{color, colortbl}
\usepackage{nicefrac}
\usepackage{bm}
\usepackage{bbding}
\usepackage{cancel}
\usepackage{bm}
\usepackage{nopageno}

\usepackage[pagebackref=true,breaklinks=true,colorlinks,bookmarks=false]{hyperref}

\def\cvprPaperID{4537} 
\def\confYear{CVPR 2021}

\begin{document}
\newcommand{\todo}[1]{\textcolor{red}{TODO: #1}}
\newcommand{\TODO}[1]{\textcolor{red}{\textbf{****** #1 ******}}}
\newcommand{\hongmin}[1]{{\color{orange}\textsc{Hongmin:} #1}}
\newcommand{\jianchao}[1]{{\color{magenta}\textsc{Jianchao:} #1}}
\newcommand{\rc}{\color{purple}}
\newcommand{\yes}{\ding{51}}
\newcommand{\no}{\ding{55}}
\def\etal{\emph{et al.}}

\title{DCNAS: Densely Connected Neural Architecture Search for Semantic Image Segmentation}

\author{Xiong Zhang$^{1}$\thanks{corresponding author, zhangxiong@yy.com}, Hongmin Xu$^2$, Hong Mo$^3$, Jianchao Tan$^2$, Cheng Yang$^1$, Lei Wang$^1$, Wenqi Ren$^4$\\
\normalsize{$^1$Joyy Inc., $^2$ AI Platform, Kwai Inc., $^3$State Key Lab of VR, Beihang University, ~$^4$SKLOIS, IIE,CAS}\\
}

\maketitle

\begin{abstract}

Existing NAS methods for dense image prediction tasks usually compromise on restricted search space or search on proxy task to meet the achievable computational demands. To allow as wide as possible network architectures and avoid the gap between realistic and proxy setting, we propose a novel Densely Connected NAS (DCNAS) framework, which directly searches the optimal network structures for the multi-scale representations of visual information, over a large-scale target dataset without proxy. Specifically, by connecting cells with each other using learnable weights, we introduce a densely connected search space to cover an abundance of mainstream network designs.
Moreover, by combining both path-level and channel-level sampling strategies, we design a fusion module and mixture layer to reduce the memory consumption of ample search space, hence favoring the proxyless searching. Compared with contemporary works, experiments reveal that the proxyless searching scheme is capable of bridging the gap between searching and training environments. Further, DCNAS achieves new state-of-the-art performances on public semantic image segmentation benchmarks, including 84.3\% on Cityscapes, and 86.9\% on PASCAL VOC 2012. We also retain leading performances when evaluating the architecture on the more challenging ADE20K and PASCAL-Context dataset.

\end{abstract}

\section{Introduction}
The heavy computation overheads of early Neural Architecture Search (NAS) methods \cite{zoph2016neural} hinder their applications in real-world problems. Recently, limited search space strategies \cite{zoph2018learning,liu2017hierarchical,liu2018darts} significantly shorten the searching time of NAS algorithms, which make NAS approaches achieve superhuman performance in image classification task.
However, to meet the low consumption of searching time, NAS methods with a constrained search space throw down the multi-scale representation of high-resolution image. 
As a result, those methods are not suitable for dense image prediction tasks (\emph{e.g.} semantic image segmentation, object detection, and monocular depth estimation).

To efficiently search appropriate network structure that can combine both the local and global clues of the semantic features, researches recently focus on improving NAS frameworks by designing new search spaces to handle multi-scale features. For instance, DPC \cite{chen2018searching} introduces a recursive search space, and Auto-DeepLab \cite{liu2019auto} proposes a hierarchical search space.
However, designing new search space for dense image prediction tasks proves challenging: one has to delicately compose a flexible search space that covers as much as possible optimal network architectures. Meanwhile, efficiently address memory consumption and the heavy computation problems accompanied with ample search space for the high-resolution imagery.

\begin{table}
\centering
\resizebox{0.99\columnwidth}{!}
{
\begin{tabular}{l  cccccc }\hline
\toprule 
\textbf{Method} & \textbf{Proxyless} &\textbf{GPU Days}& \textbf{FLOPs(G)}& $\bm{\rho \uparrow}$ & \bm{$\tau \uparrow$} & \textbf{mIoU(\%)} \bm{$\uparrow$}\\
\toprule 
DPC \cite{chen2018searching} & \no & 2600 &684.0 & 0.46 & 0.37 & 82.7 \\
Auto-DeepLab \cite{liu2019auto} & \no &3 &695.0& 0.31 & 0.21  & 82.1 \\
CAS \cite{zhang2019customizable} & - & - &- & - & -  & 72.3 \\
GAS \cite{lin2020graph} & - &6.7 & - &-  &-  & 73.5 \\
FasterSeg \cite{chen2019fasterseg} &\no& 2 & 28.2 &0.35 & 0.25 & 71.5 \\
Fast-NAS \cite{nekrasov2019fast} &\no& 8 & 435.7&0.42 & 0.33  & 78.9 \\
SparseMask \cite{wu2019sparsemask} &\no& 4.2 &36.4& 0.49 & 0.38  & 68.6 \\
\hdashline
Ours (DCNAS) &\yes& 5.6 & 294.6 &0.73 & 0.55 & 84.3 \\
\bottomrule
\end{tabular}
}
\vspace{3pt}
\caption{\textbf{Comparisons with state of the arts.} 
The table  presents the comparison results of \cite{chen2018searching,liu2019auto,zhang2019customizable,lin2020graph,chen2019fasterseg,li2020learning,wu2019sparsemask} and our DCNAS comprehensively. In which, \bm{$\rho$} and \bm{$\tau$} represent the Pearson Correlation Coefficient and the Kendall Rank Correlation Coefficient, respectively, both of which are measured according to  the performances of the super-net and the fine-tuned stand-alone model, \textbf{Proxyless} applies a proxyless searching paradigm,  \textbf{GPU Days} presents the searching cost, \textbf{mIoU(\%)} and \textbf{FLOPs(G)} report the accuracy and the flops of the best model.
}
\end{table}

\begin{figure*}[ht]
\centering
\includegraphics[width=0.9\textwidth]{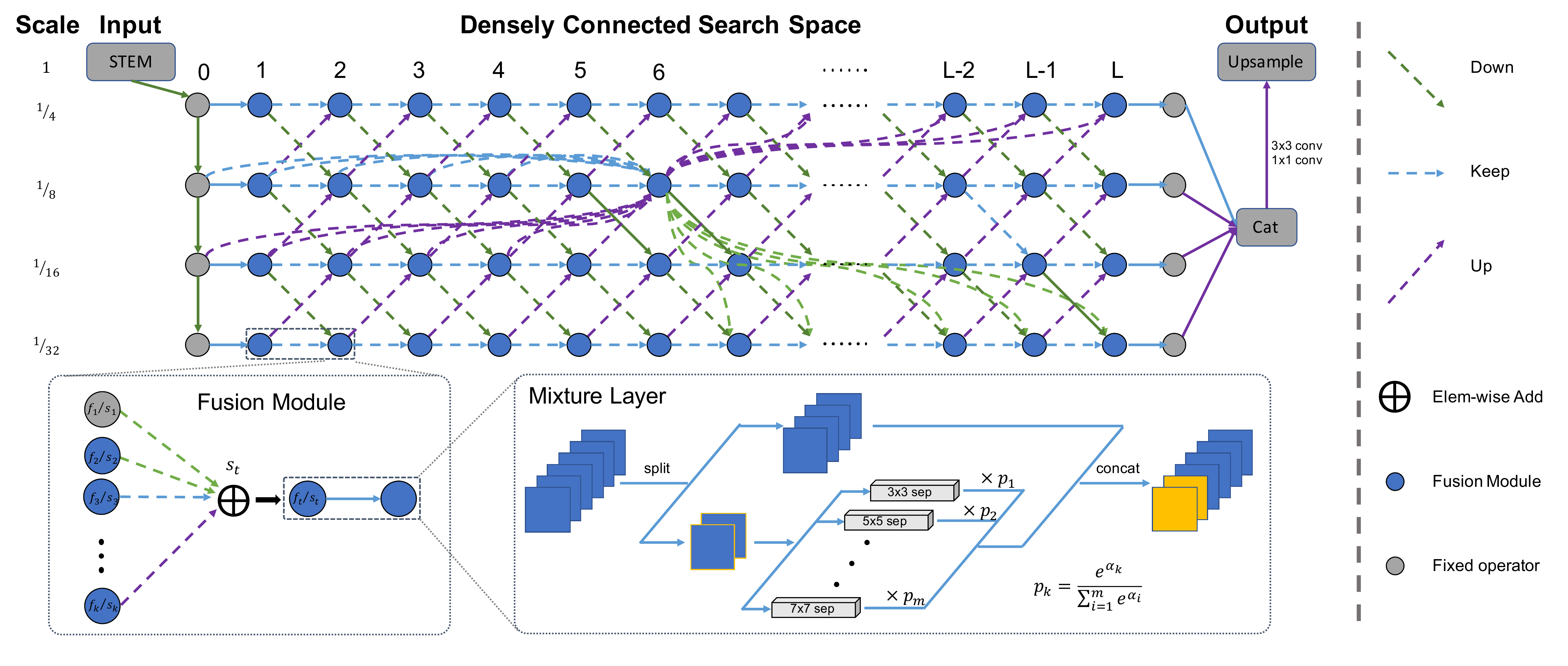}
\caption{\textbf{Framework of DCNAS.} 
\emph{Top:}  The \emph{densely connected} search space (\textbf{DCSS}) with layer $L$ and max downsampling rate 32.
To stable the searching procedure, we keep the beginning \emph{STEM} and final \emph{Upsample} block unchanged.
\emph{Dashed lines} represent candidate connections in DCSS, to keep clarity, we only demonstrate several connections among all the candidates.
\emph{Bottom Left:}  \emph{Fusion module} targets at aggregating feature-maps derived by previous layers, and solving the intensive GPU memory requirement problem by sampling a portion of connections from all possible ones.
\emph{Bottom Right:} \emph{Mixture layer} may further save GPU memory consumption and accelerating the searching process by sampling and operating on a portion of features while bypassing the others.
}

\label{fig:main_framework}
\end{figure*}

In this work, we propose an efficient and proxyless NAS framework to search the optimal model structure for semantic image segmentation.
Our approach is based on two principal considerations.
Firstly, the search space should be comprehensive enough to handle most of the mainstream architecture designs, even some undiscovered high-quality model structures.
As shown in Figure  \ref{fig:main_framework}, we design a reticular-like and fully densely connected search space, which contains various paths in the search space. Consequently, DCNAS may derive whichever model architecture designs by selecting appropriate paths among the whole set of these connections.
Additionally, DCNAS aggregates contextual semantics encoded in multi-scale imageries to derive long-range context information, which has been recently found to be vital in the state-of-the-art manually designed image segmentation models \cite{liu2015parsenet,zhao2017pyramid,chen2017rethinking,chen2018encoder,zhang2019co}.
Secondly, we observe that previous NAS approaches \cite{chen2018searching,liu2019auto} heavily rely on the proxy task or proxy dataset to reduce the cost of GPU hours and to alleviate the high GPU memory consumption problem.
However, architectures optimized over the proxy are not guaranteed to be suitable in realistic setting \cite{cai2018proxylessnas}, because of the gap between the proxy and the target configuration.
Regarding this concern, we relax the discrete architectures into continuous representation to save GPU hours and design a fusion module which applies both path-level and channel-level sampling strategies during the searching procedure to reduce memory demand. 
Based on that, one may employ the stochastic gradient descent (SGD) to perform the proxyless searching procedure to select the optimal architecture from all the candidate models without the help of proxy datasets or proxy tasks. The searching procedure takes about 5.6 GPU days on Cityscapes \cite{cordts2016cityscapes}.

We apply our approach to semantic image segmentation tasks on several public benchmarks, our model  achieves the best performance compared with state-of-the-art hand-craft models \cite{rota2018place,zhao2017pyramid,zhuang2018dense,chen2018encoder,huang2019ccnet,ding2019semantic,fu2019dual} and other contemporary NAS approaches \cite{liu2019auto,chen2018searching,wu2019sparsemask,nekrasov2019fast,chen2019fasterseg}.
We also evaluate the optimal model identified by DCNAS on PASCAL VOC 2012 \cite{everingham2010pascal}, the model outperforms other leading approaches \cite{ke2018adaptive,fu2019dual,zhao2017pyramid,yu2018learning,zhang2018context,li2018pyramid,zhang2019co,he2019dynamic} and advances the state-of-the-art performance.
Transferring the model to ADE20K \cite{zhou2017scene} and  PASCAL-Context \cite{mottaghi2014role} datasets, our model obtains the best result compared with state-of-the-art approaches \cite{ding2019semantic,fu2019dual,he2019adaptive,huang2019ccnet,zhang2017scale,zhang2018context,liang2018dynamic}, according to the corresponding evaluation metrics.

Further,  considering NAS methods depend on agent metrics to explore promising networks efficiently. 
If the agent metric provides a strong relative ranking of models, this enables the discovery of high performing models when trained to convergence.
However, studies about the relative ranking of models in the searching period are less explored in literature \cite{nekrasov2019fast,wu2019sparsemask,chen2019fasterseg,chen2018searching,liu2019auto}.
In this work, we enrich this field and investigate most related methods on how well the accuracy in searching phase
for specific architectures correlates with that of the fine-tuned stand-alone model.

To summarize, our main contributions are as follows:
\begin{enumerate}[$\bullet$]
\item We design a novel \emph{entirely densely connected search space}, allowing to explore various existing designs and to cover arbitrary model architecture patterns. 
\item A novel \emph{proxyless searching paradigm} is implemented to efficiently and directly explore the most promising model among all the candidates encoded in DCNAS, on large-scale segmentation datasets (\emph{e.g.,} Cityscapes \cite{cordts2016cityscapes}).
\item The DCNAS outperforms  contemporary NAS methods \cite{nekrasov2019fast,wu2019sparsemask,chen2019fasterseg,chen2018searching,liu2019auto} and demonstrates new state-of-the-art performance on Cityscapes \cite{cordts2016cityscapes}, PASCAL-VOC 2012 \cite{everingham2010pascal}, PASCAL-Context  \cite{mottaghi2014role} and ADE20K \cite{zhou2017scene} datasets.
\item We conduct a large scale experiment to understand the correlation of model performance between the searching and training periods for most contemporary works  \cite{liu2019auto,chen2018searching,wu2019sparsemask,nekrasov2019fast,chen2019fasterseg} focusing on semantic segmentation task.
\end{enumerate}

\section{Related Work}
\subsection{Neural Architecture Search Space}
Early NAS works \cite{zoph2016neural,real2017large} directly search the whole network architecture. Although those works achieve impressive results, their expensive computation overheads (\emph{e.g.} thousands of GPU days) hinder their applications in real-world scenarios. To alleviate this situation, researchers proposed restricted search spaces. 
NASNet \cite{zoph2018learning} first proposed a cell-based search space.
Concretely, NASet searches the fundamental cell structure, which is easier than searches the whole network architectures on a small proxy dataset. 
After that, many works \cite{liu2018darts,pham2018efficient,cai2018proxylessnas,tan2019mnasnet,wu2019fbnet} adopt the cell-based search space design and make improvements in several ways. One line of works \cite{cai2018proxylessnas,tan2019mnasnet,wu2019fbnet} attempt to search high-efficiency network architectures for resource-constrained platforms, such as mobile phones. 
Another line of works \cite{chen2019progressive,fang2019densely} try to improve previous cell-based NAS methods. P-DRATS \cite{chen2019progressive} first bridge the depth gap between search and evaluation phases. Fang \etal \cite{fang2019densely} proposes the routing blocks to determine the number of blocks automatically. 

Despite the success of repeated cell-based methods for image classification tasks, other dense image prediction tasks, such as semantic image segmentation and object detection, demand more delicate and complicated network architectures to capture multi-scale information of the image. 
Therefore, researchers begin to explore more flexible search space for dense image prediction tasks.
For instance, DPC \cite{chen2018searching} proposes a recursive search space to build a multi-scale representation of a high-resolution image. Auto-DeepLab \cite{liu2019auto} introduces a hierarchical search space to exhibit multi-level representations of the image.

\subsection{Semantic Image Segmentation}
The pioneer work FCN \cite{long2015fully} makes great progress in semantic segmentation with a fully convolutional network. Afterwards, researchers devote enormous efforts to explore the tremendous potentials of convolutional neural networks (CNNs) on dense image prediction tasks. Several kinds of research works \cite{zhao2017pyramid,lin2016efficient,chen2016attention,eigen2015predicting,chen2017deeplab,ronneberger2015u,newell2016stacked,lin2017refinenet,pohlen2017full,chen2018encoder,fu2019stacked} focus on how to efficiently utilize multi-scale context information to improve the performance of semantic segmentation systems. Some works \cite{lin2016efficient,chen2016attention,eigen2015predicting} takes an image pyramid as input to capture larger objects in down-sampled input image. Some other works try to make use of multi-scale context information. PSPNet \cite{zhao2017pyramid} uses spatial pyramid pooling at different manually set scales. Deeplab \cite{chen2017deeplab} performs multiple rates atrous convolution operations. Also, there are some works \cite{ronneberger2015u,newell2016stacked,lin2017refinenet,pohlen2017full,chen2018encoder,fu2019stacked} that employ encoder-decoder structure to captures long-range context information.

Recent works \cite{chen2018searching,liu2019auto,nekrasov2019fast,shaw2019squeezenas,wu2019sparsemask,chen2019fasterseg,li2019partial,zhang2019customizable,zhang2019end,wu2020auto,liu2020scam} shift to automatically design network architectures. Research works, such as \cite{nekrasov2019fast,shaw2019squeezenas,wu2019sparsemask,chen2019fasterseg,li2019partial,zhang2019customizable}, target to search network structures that are friendly to resource-constrained devices for semantic segmentation task. 
Other works, such as DPC \cite{chen2018searching} and Auto-DeepLab  \cite{liu2019auto}, intent to directly search the highest-performance network architecture for semantic segmentation without considering computation overheads. 
Most of those works \cite{liu2019auto,nekrasov2019fast,shaw2019squeezenas,wu2019sparsemask,chen2019fasterseg,li2019partial,zhang2019customizable} suffer from limited search space and search on proxy tasks, because of high GPU memory consumption and expensive computational burden. Yet the optimal network structures for multi-scale feature representations require ample search space. 

The most similar works to ours are DPC \cite{chen2018searching}, Auto-DeepLab \cite{liu2019auto}, and DenseNAS \cite{fang2019densely}. 
~\cite{chen2018searching} proposes an ample search space but requires thousands of GPU days for searching, whereas we can efficiently search over at least the same search space using only 5.6 GPU days. 
~\cite{liu2019auto} achieves comparable searching time like ours, yet contains a much smaller search space than DCNAS.
~\cite{fang2019densely} places candidate connections between adjacent $M$ cells only, while DCNAS comprises a wholly densely connected search space.
Besides, both \cite{chen2018searching,liu2019auto,fang2019densely} search on a proxy dataset, whereas DCNAS directly search on the target dataset.




\section{Methodology}
In this section, we introduce several vital components that constitute DCNAS in detail, including a densely connected search space that is general enough to cover enormous network design, the derived continuously differentiable representation of the search space, and the approach that can save the cost of GPU hours and reduce excessive memory footprint, which favors proxyless searching.





\begin{figure}[ht]
	\centering
	\subfigure[Network Architecture of FCN-32s \cite{long2015fully}]{\label{fig:dcnas_fcn} \includegraphics[width=0.9\columnwidth]{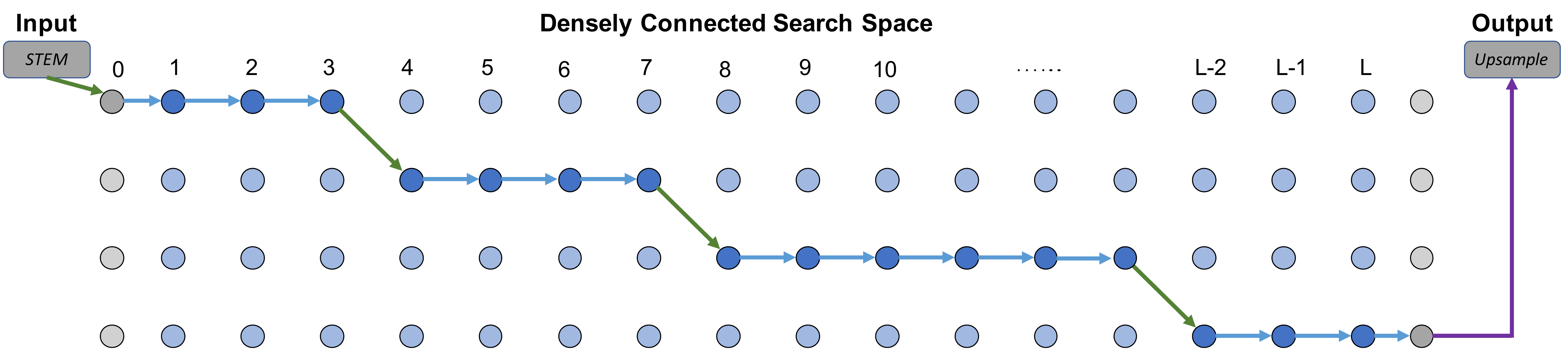}}	
	\subfigure[Network Architecture of HourGlass \cite{newell2016stacked}]{\label{fig:hourglass} \includegraphics[width=0.9\columnwidth]{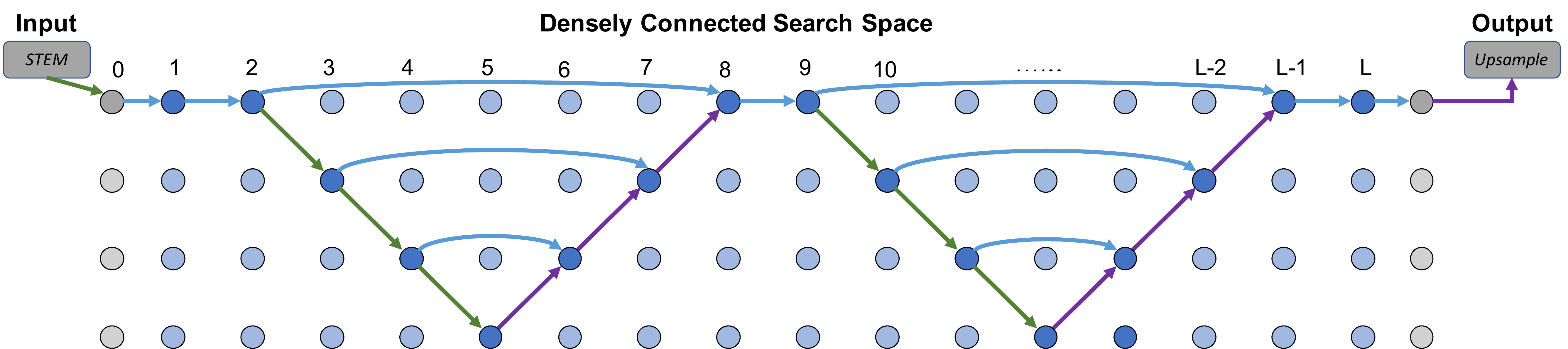}}	
	\subfigure[Network Architecture of DeepLabV3 \cite{chen2017rethinking}]{\label{fig:dcnas_deeplabv3} \includegraphics[width=0.9\columnwidth]{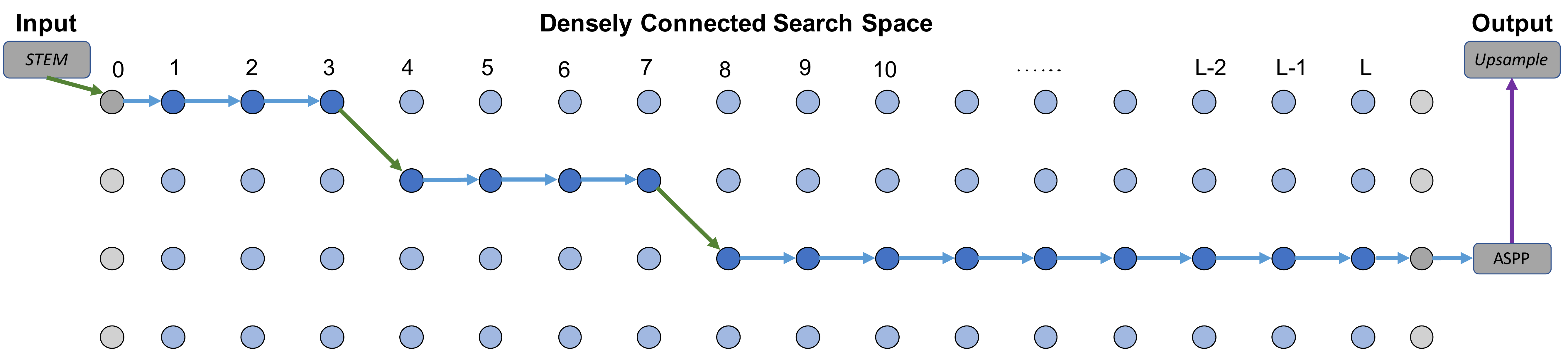}}	
	\subfigure[Network Architecture of HRNet \cite{sun2019deep}]{\label{fig:dcnas_hrnet} \includegraphics[width=0.9\columnwidth]{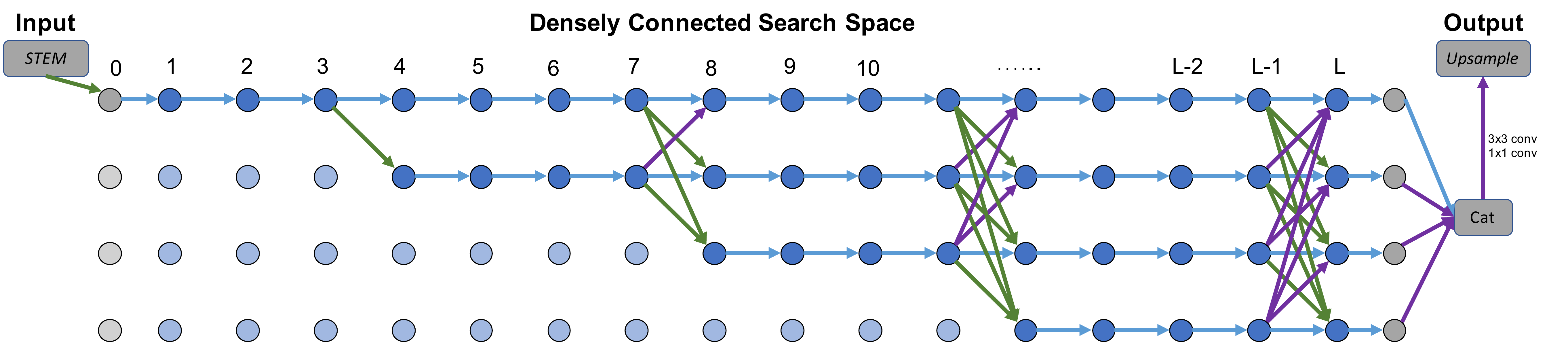}}	
	\caption{\textbf{Mainstream Network Designs in DCNAS.} 
	The figure demonstrates several mainstream network design patterns \cite{long2015fully,newell2016stacked,chen2017rethinking,sun2019deep} that embedded in our \emph{Densely Connected Search Space}.
}
\label{fig:embedded}
\end{figure}

\subsection{Densely Connected Search Space}
The densely connected search space ( \textbf{DCSS} ) involves two primitives, the  \emph{mixture layer}, and the \emph{fusion module}.
The \emph{mixture layer} is defined as a mixture of the candidate operators, and the \emph{fusion module} aims at aggregating semantic features from preceding fusion modules.
A range of reticular-like and densely connected fusion modules constitute the architecture of DCSS, as shown in Figure~\ref{fig:main_framework}.

\textbf{Mixture Layer.}  
The \emph{mixture layer} is the elementary structure in search space represents a collection of available operations. 
Similar to \cite{cai2018proxylessnas}, we construct the operator space $\mathcal{O}$ with various configurations of mobileNet-v3 \cite{howard2019searching},
\emph{i.e.,}  kernel sizes $k \in\{{3,5,7}\}$, expansion ratios $r \in \{{3,6}\}$. 
Due to the dense connections contained in the search space, the DCSS inherently supports the zero operation and identity mapping. Hence $\mathcal{O}$ does not contain the above operators explicitly.
Similarly, considering the search space inherently favors the multi-scale semantic features aggregation, which increases the receptive field, therefore, the mixture layer does not support atrous convolution explicitly.

\textbf{Fusion Module.} 
To explore various paths in DCSS, we introduce the \emph{fusion module} with the ability of aggregating semantic features from preceding fusion modules besides the feature pyramids and attaching transformed semantic features to succeeding ones.
As illustrated by the bottom-left part in Figure \ref{fig:main_framework}, the fusion module consists of the shape-alignment layer besides the mixture layer.
The shape-alignment layer is in a multi-branch parallel form.
Given an array of feature-maps with different shapes (\emph{e.g.,} spatial resolutions and channel widths), the shape-alignment layer dispatches each feature-map to the corresponding branches to align them to the target shape.
Semantic features are well-aligned and fully aggregated, then feed into the mixture layer to perform efficient multi-scale features fusion.

\textbf{Search Space.}
Benefiting from the sophisticated design of the fusion module, we introduce a set of reticular-arranged fusion modules and place dense connections between them to form a search space that supports more flexible architecture configurations than similar works \cite{liu2019auto,chen2019fasterseg}.
The search algorithm is admitted to select a subset of the fusion models together with the appropriate connections between them to derive whatever model architecture  (see Figure \ref{fig:embedded} for instance), hence enables a general architecture search.
Specifically, we construct the super network with an array of fusion modules $\{\mathcal{M}_{(s,l)} \}$, 
where $s$ indicates the spatial configuration and $l$ refers to the layer index, which is demonstrated in Figure \ref{fig:main_framework},
in which, each $\mathcal{M}_{(s,l)}$ aggregates the semantic features coming from $\{\mathcal{M}_{(s^{\prime}, l^{\prime})}\}_{l^\prime <l}$. 

\subsection{Differentiable Representation}
Reinforcement learning representations in \cite{baker2016designing,zoph2016neural} and evolutionary representations in \cite{suganuma2017genetic,real2017large} both tend to be computationally intensive, hence probably not suitable for semantic image segmentation task. 
Regarding this concern, we draw on the experience of continuous relaxation in \cite{liu2018darts,xie2018snas} and relax the discrete architectures into continuous representation.
Notably, we derive a continuous representation for both the mixture layer and the fusion module, hence leading to a fully differentiable search space, based on which one can apply the stochastic gradient descent method to search promising architectures.

\textbf{Mixture Layer}. 
We assign an architecture parameter $\alpha_{(s,l)}^o$ to each operator $o$ in mixture layer $\ell_{(s,l)}$ contained in fusion module $\mathcal{M}_{(s,l)}$, and derive the continuously representation of the mixture layer by defining it as a weighted sum of outputs from all candidate operations.
The architecture weight of the operation is computed as a soft-max over all architecture parameters in mixture layer $\ell_{(s,l)}$,
\begin{align}
\label{eq:mix_w}
\tiny{
w_{(s,l)}^{o}=\frac{\exp \{ \alpha_{(s,l)}^o \}}{\sum_{o^{\prime} \in \mathcal{O}} \exp \{ \alpha_{(s,l)}^{o^\prime} \}},
}
\end{align}
and the output of mixture layer $\ell_{(s,l)}$ can be estimated as,
\begin{align}
\label{eq:mix_full}
O_{(s,l)}=\sum_{o \in \mathcal{O}} w_{(s,l)}^{o}  o\left(\mathcal{I}_{(s,l)}\right),
\end{align}
where $\mathcal{I}_{(s,l)}$ refers to the input feature-maps of  $\ell_{(s,l)}$.

\textbf{Fusion Module.}
Similar to the relaxation paradigm adopted in the mixture layer, to relax the connections as a continuous representation, we assign an architecture parameter $\beta_{(s^\prime,l^\prime)\rightarrow(s,l)}$ for the path from $\mathcal{M}_{(s^\prime,l^\prime)}$ to $\mathcal{M}_{(s,l)}$.
Since the $\mathcal{M}_{(s,l)}$ aggregates all semantic features $\{\mathcal{I}_{(s^\prime,l^\prime)}\}_{l^\prime< l}$, we estimate the transmission probability of each path using a soft-max function over all available connections,
\begin{align}
\label{eq:fm_prob}
\small{
p_{(s^\prime,l^\prime)\rightarrow(s,l)}=\frac{\exp{\left \{ \beta_{(s^\prime,l^\prime)\rightarrow(s,l)} \right \}}}{ \sum_{s^{\prime\prime} \in \mathbb{S}, \, l^{\prime\prime}   \,<\, l} \exp{\left\{  \beta_{(s^{\prime\prime},l^{\prime\prime})\rightarrow(s,l)} \right\}}},
}
\end{align}
and the aggregating process can be formalized as,
\begin{align}
\label{dq:calc_fm_full}
\tiny{
\mathcal{I}_{(s,l)}=\sum_{s^\prime \in \mathbb{S}, l^\prime < l}p_{(s^\prime,l^\prime)\rightarrow(s,l)} H_{(s^\prime,l^\prime)\rightarrow(s,l)}(O_{(s^\prime,l^\prime)}),
}
\end{align}
where $H_{(s^\prime,l^\prime\rightarrow(s,l)}$ denotes the corresponding transformation branch in shape-alignment layer of fusion module $\mathcal{M}_{(s,l)}$, which transforms the semantic features produced by mixture layer $\ell_{(s^\prime,l^\prime)}$ to the shape specified by $\mathcal{M}_{(s,l)}$.
\subsection{Reducing Memory Footprint}
Continuous representation of network architectures can largely reduce the cost of GPU hours, while the memory footprint grows linearly w.r.t. the size of candidate operation set and the number of connections, hence suffering from the high GPU memory consumption problem.
While image segmentation intrinsically requires high-resolution semantic features, which may lead to even more GPU memory consumption.
To resolve the excessive memory consumption problem, 
we apply the sampling strategy both in the fusion module and the mixture layer to solve the excessive memory footprint problem and further reduce the cost of GPU hours.
Taking the above tactics, one can proxylessly search the model over large-scale dataset, hence avoiding the gap between target and proxy dataset.

%
%
%
%

\textbf{Fusion Module.} The DCSS comprises $8L(L+1)$ connections in total, where $L$ is layer number, thus making it impossible to optimize the whole set of candidate connections in each iteration because of the extravagant GPU memory demand.
Therefore, in each iteration during the search, for each fusion module, we sample several connections among the corresponding potential connections.
Concretely, taking fusion module $\mathcal{M}_{(s,l)}$ for instance, all the preceding fusion modules $\{\mathcal{M}_{(s^\prime,l^\prime)}\}_{l^\prime< l}$ are associated with it, in each search iteration, we perform sampling without replacement to activate several transmission paths, and the probability distribution is defined as:
\begin{align}
\tiny{
p_{(s^\prime,l^\prime)\rightarrow(s,l)}=\frac{\exp{\left \{ \beta_{(s^\prime,l^\prime)\rightarrow(s,l)} /\tau \right \}}}
{ \sum_{s^{\prime\prime}\in \mathbb{S}, l^{\prime\prime} < l} \exp{\left\{  \beta_{(s^{\prime\prime},l^{\prime\prime})\rightarrow(s,l)} / \tau\right\}}},
}
\end{align}
where $  \beta_{(s^{\prime\prime},l^{\prime\prime})\rightarrow(s,l)}$ share the same definition in Equation \ref{eq:fm_prob}, and the temperature variable $\tau$ starts from a high temperature then anneal to a small but non-zero value. 
Assume there are $n$ fusion modules $\{ \mathcal{M}_{s_i,l_i} \}_{i < n}$ selected from  $\{\mathcal{M}_{(s^\prime,l^\prime)}\}_{l^\prime< l}$, 
then we may approximate Equation \ref{dq:calc_fm_full} with:
\begin{align}
\mathcal{I}_{(s,l)}=\sum_{k < n}w_{(s_k,l_k)\rightarrow(s,l)} H_{(s_k,l_k)\rightarrow(s,l)}(O_{(s_k,l_k)}),
\end{align}
where $w_{(s_k,l_k)\rightarrow(s,l)}$ refers to the normalized blending weight, $H$ and $O$ shares the same definition in Equation \ref{dq:calc_fm_full}.

\textbf{Mixture Layer.} 
A typical way to apply the sampling trick is to sample the operator $o$ from the candidate operator space $\mathcal{O}$ in each mixture layer.
While in practice, we find that the naive sampling strategy makes the searching process unstable and sometime does not convergence.
Inspired by \cite{xu2019pc}, a portion of channels are sampled among the input features, then the lucky ones are sent into the mixed transformation of  $|\mathcal{O}|$ operators while bypasses the others. 
Specifically, for each $\ell_{(s,l)}$,  we assign a random variable $S_{(s,l)}$ that control the ratio of sampled channels, which trades off the search accuracy and efficiency, and we may estimate $O_{(s,l)}$ in Equation \ref{eq:mix_full} as:
\begin{align}
\tiny{
\sum_{o\in \mathcal{O}} w_{(s,l)}^o  o \left(S_{(s,l)} \mathcal{I}_{(s,l)}\right) + (1-S_{(s,l)}) \mathcal{I}_{(s,l)},
}
\end{align}
where $w_{(s,l)}^o$ indicates the normalized blending weight defined by Equation \ref{eq:mix_w}.
While different to \cite{xu2019pc}, the edge normalization layers are removed and the feature masks $S_{(s,l)}$ remain unchanged in the searching procedure, and experimental result demonstrates that this simplified variant proves more suitable for semantic image segmentation.
Regarding the sampling ratio, we do not tune the sampling ratio and fix it to be $\nicefrac{1}{4}$, empirically, our experiments demonstrate that it works well. With this scheme, one shall gain $4 \,\times$ acceleration of the searching process, and save about $75\%$ GPU memory consumption.
\begin{figure*}[ht]
	\centering
	\subfigure[Fast-NAS \cite{nekrasov2019fast} $\rho$: 0.42, $\tau$: 0.33]{\label{fig:fast_nas}\includegraphics[width=0.3\textwidth]{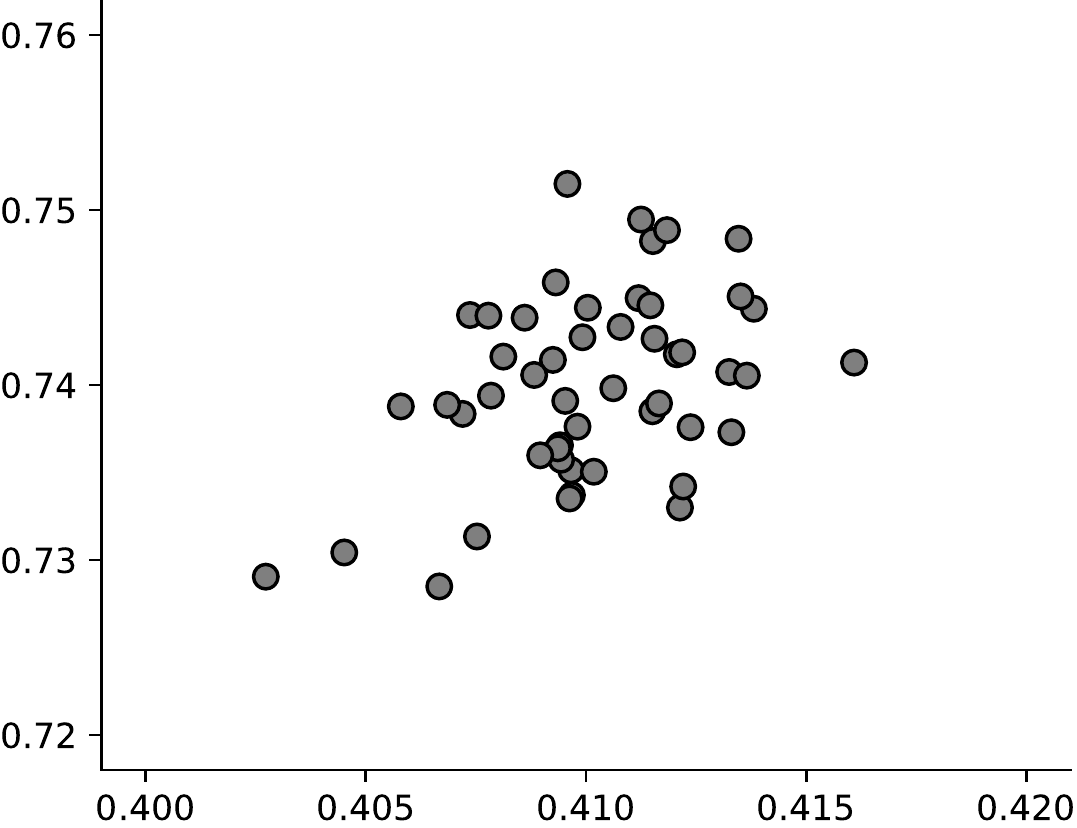}}
	\subfigure[Sparse-Mask \cite{wu2019sparsemask} $\rho$: 0.49, $\tau$: 0.38]{\label{fig:sparse_mask}\includegraphics[width=0.3\textwidth]{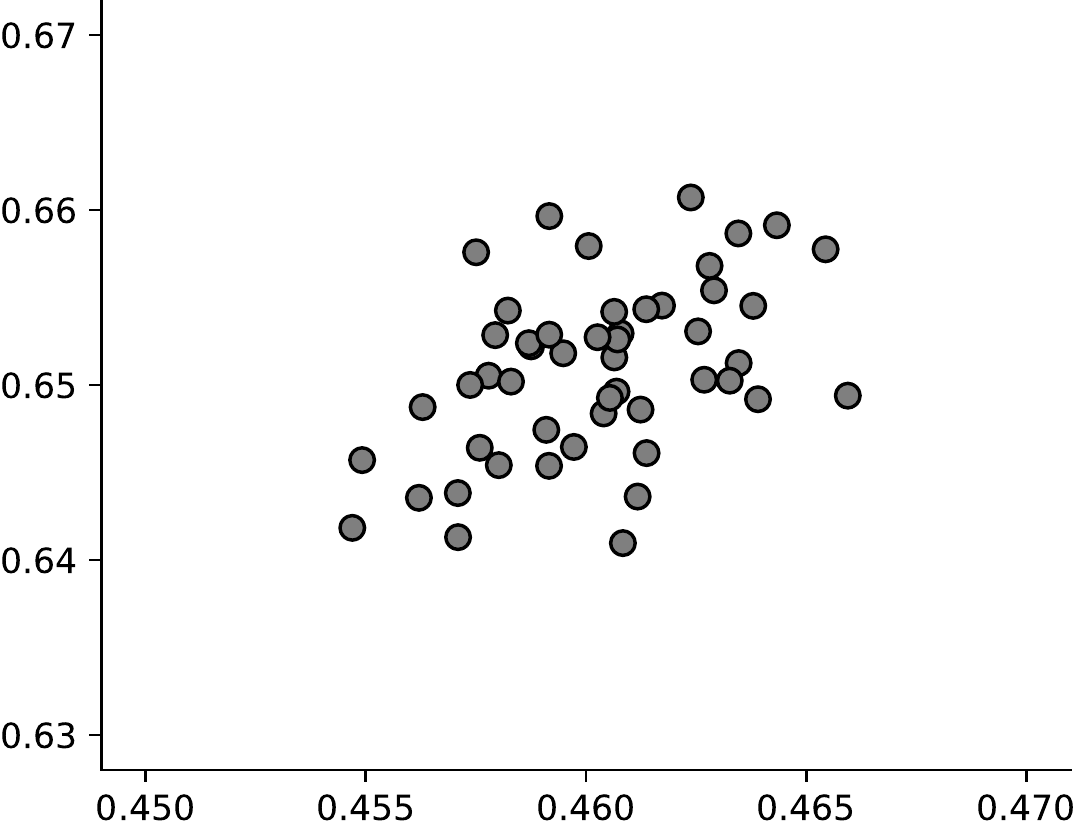}}
	\subfigure[FasterSeg \cite{chen2019fasterseg} $\rho$: 0.35, $\tau$: 0.25]{\label{fig:Faster_seg}\includegraphics[width=0.3\textwidth]{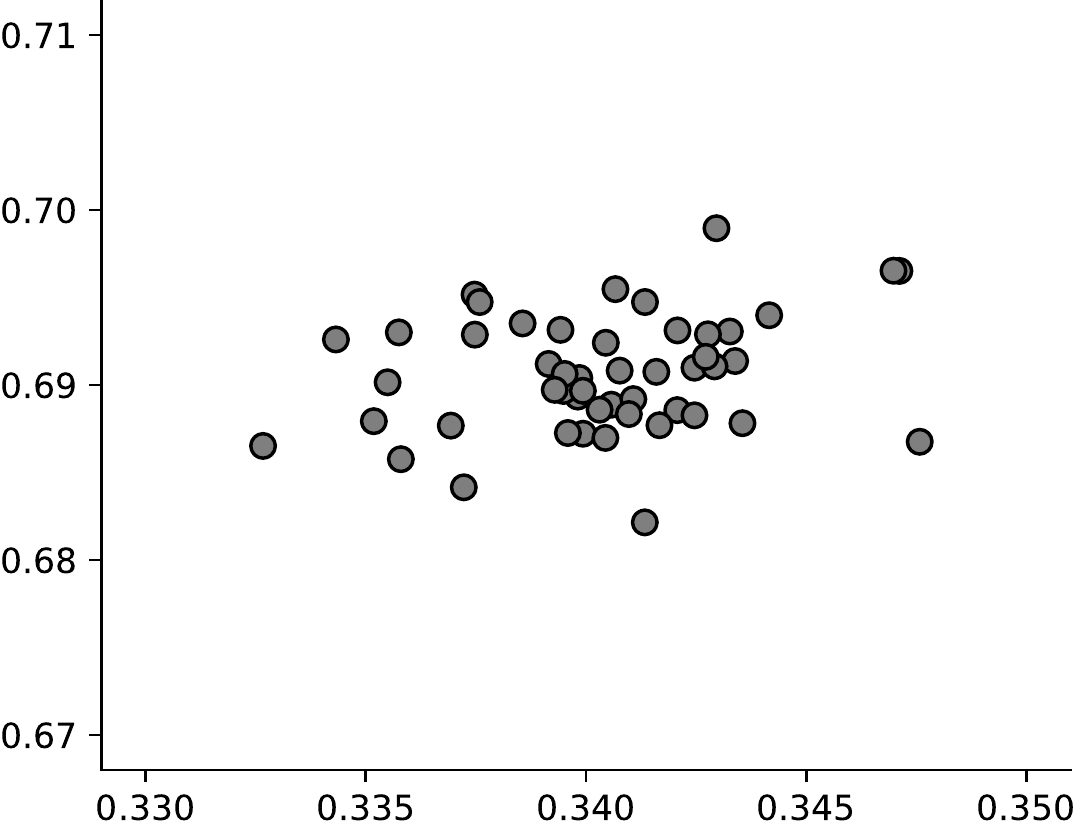}}
	\subfigure[DPC \cite{chen2018searching} $\rho$: 0.46 $\tau$: 0.37]{\label{fig:dpc}\includegraphics[width=0.3\textwidth]{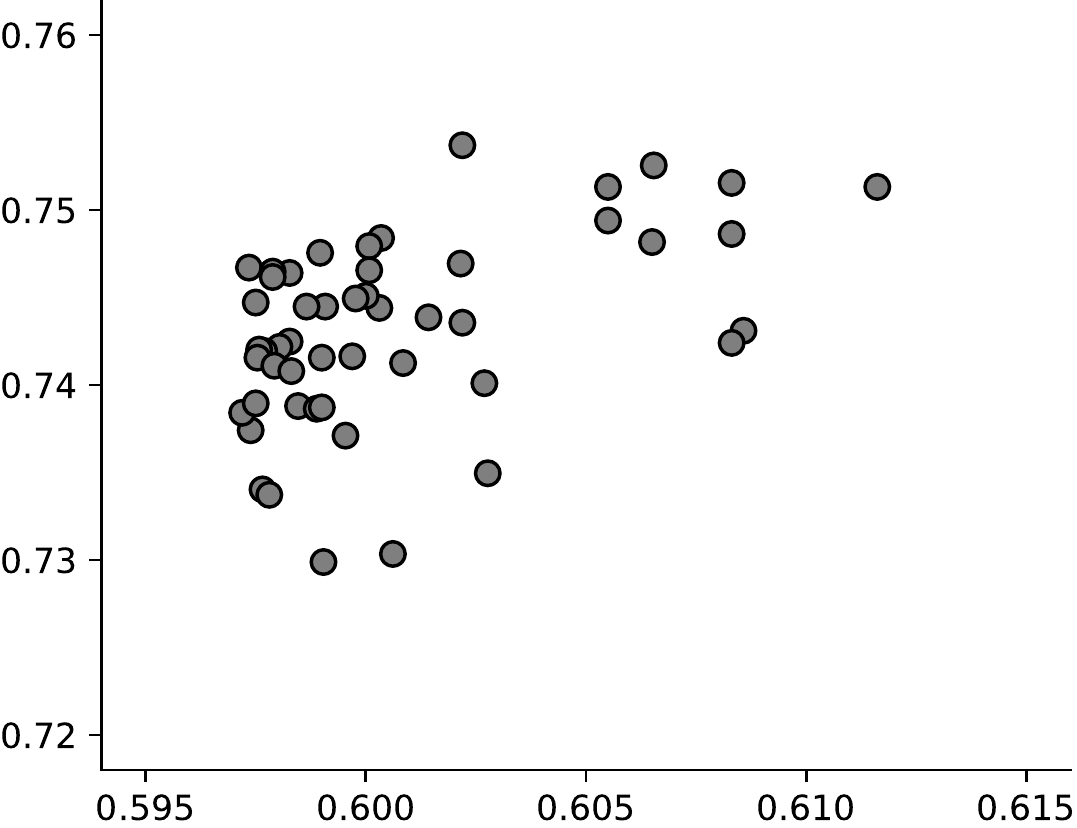}}
	\subfigure[Auto-Deeplab \cite{liu2019auto} $\rho$: 0.31, $\tau$: 0.21]{\label{fig:auto_deeplab}\includegraphics[width=0.3\textwidth]{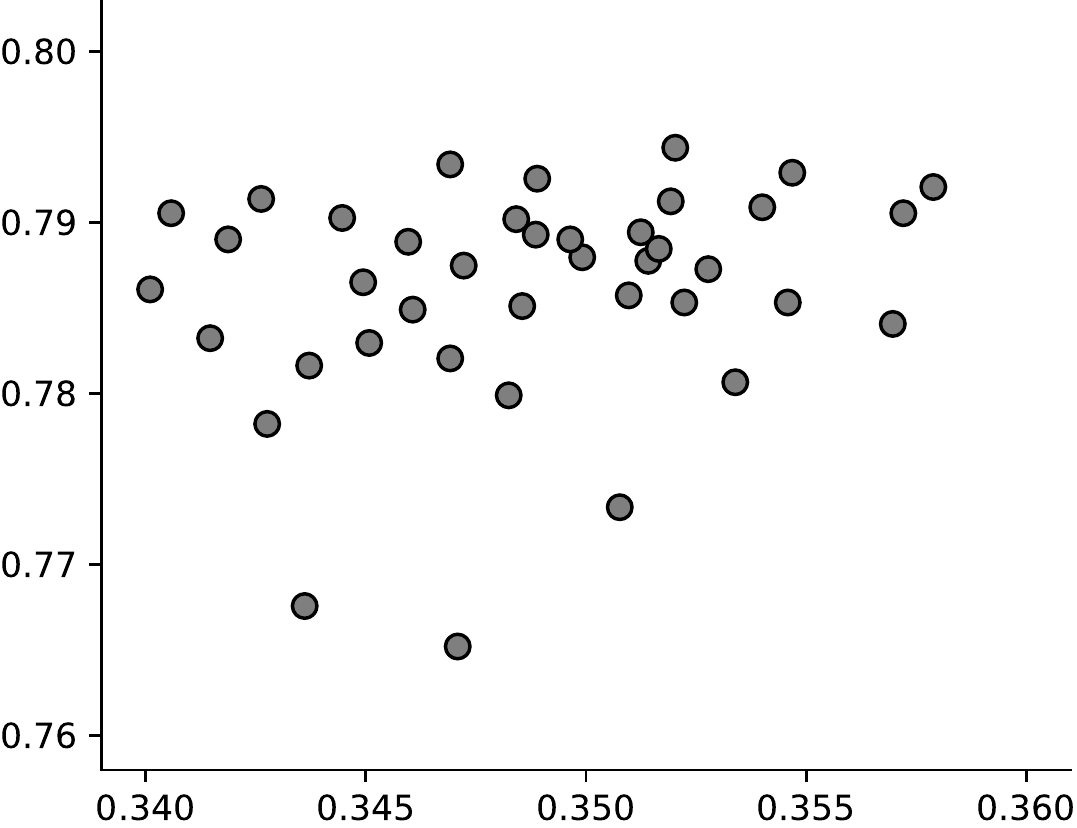}}	
	\subfigure[Ours (DCNAS) $\rho$: 0.73, $\tau$: 0.55]{\label{fig:dcnas}\includegraphics[width=0.3\textwidth]{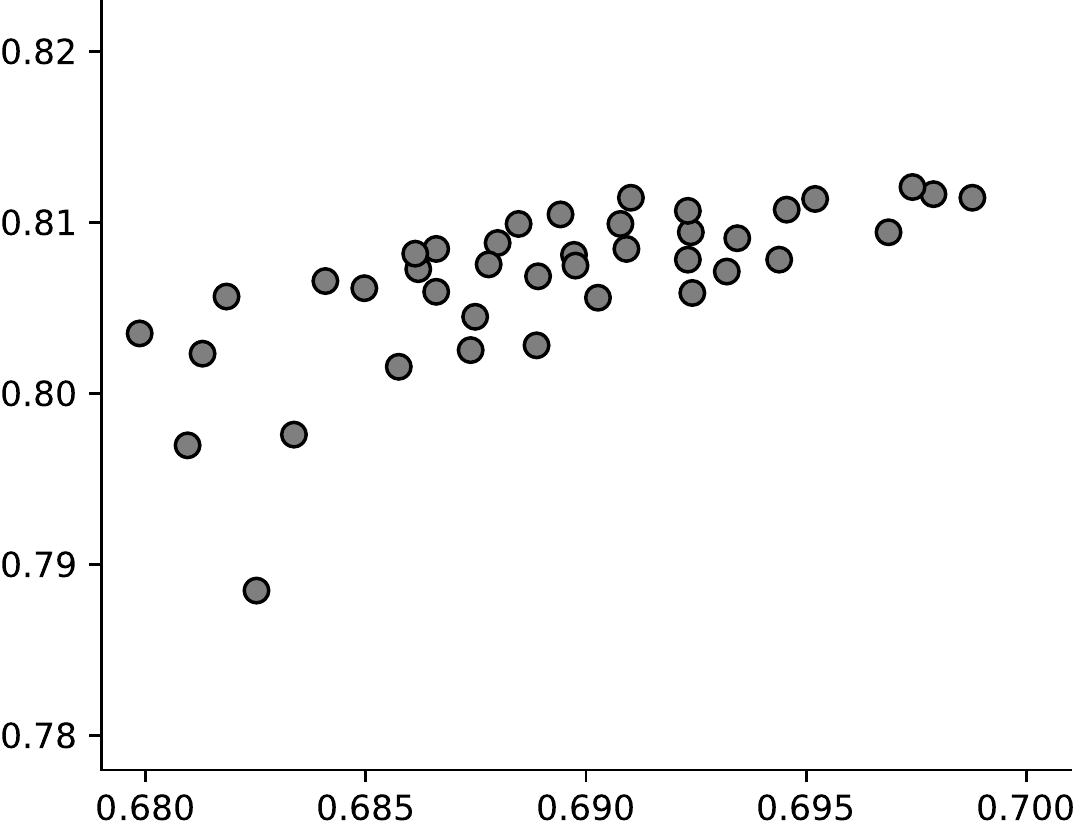}}
	\caption{\textbf{Correlation of Performance.} 
	The figures present the performance of the state of the art methods \cite{nekrasov2019fast,wu2019sparsemask,chen2019fasterseg,chen2018searching,liu2019auto} and our DCNAS.
For each method, we plot the validation performance for both the searching and training stages. 
Further, we report the Pearson Correlation Coefficient ($\rho$) and the Kendall Rank Correlation Coefficient ($\tau$), which measured according to pairs of the performances. 
\label{fig:coef}
}
\end{figure*}

\subsection{Search Procedure}
Benefiting from the fully differentiable representation of the search space and the sampling strategies mentioned above, one may apply the stochastic gradient descent algorithm to search for appropriate models by optimizing architecture parameters $\{\alpha,\beta\}$ on the target dataset without proxy.
Similar to previous works \cite{liu2018darts,chen2018searching,liu2019auto,chen2019fasterseg}, we partition the training data into two parts \emph{trainA} and \emph{trainB}, which are used for updating the convolutional weights $w$ and architecture parameters $\{\alpha,\beta\}$ respectively.
We solve this optimization problem alternatively:
\begin{enumerate}[$\bullet$]
  \item update weights $w$ by $\nabla_{w} \, \mathcal{L}_{trainA}(w,\alpha,\beta)$,
  \item update parameters $\{\alpha,\beta\}$ by $\nabla_{\alpha,\beta} \, \mathcal{L}_{trainB}(w,\alpha,\beta)$,
\end{enumerate}
in which the loss function $\mathcal{L}$ mainly consists of the classical cross entropy calculated on each mini-batch.

We shall point out that, in practice, $\mathcal{L}$ comprises of several practical regularization terms in addition to cross-entropy, experiments reveal the regularization terms yield a faster convergence rate of the searching procedure and lead to a better searching accuracy (see supplementary material).



\subsection{Decoding Network Structure}
Once the search procedure terminates, one may derive the suitable operator for each mixture layer and the optimal architecture based on the architecture parameters $\alpha$ and $\beta$. 
For mixture layer $\ell_{(s,l)}$, we select the candidate operation that has maximum operation weight, \emph{i.e.,} 
$\mathop{\arg\max}_{o\,\in \,\mathcal{O}} \alpha^o_{(s,l)}$.
In terms of the network architecture, we use Breadth-First Search algorithm (see supplemental material) to derive the network architecture.

\section{Experiments}
To evaluate DCNAS comprehensively, we first verify the effect of the proxyless searching paradigm by measuring the correlation of the performance between the searching and training environments.
Secondly, we quantitatively demonstrate the superiority of DCNAS on several broadly used benchmarks according to corresponding evaluation metrics.
Finally, we perform ablation studies to better understand the impact of different designs on the segmentation tasks.


\subsection{Architecture Search Implementation Details}
We first search suitable model structure on Cityscapes \cite{cordts2016cityscapes} dataset and then evaluate this derived model on other benchmarks.
In our experiments, for the shape of feature-maps, we set the spatial resolution space $\mathbb{S}$ to be $\{\nicefrac{1}{4},$ $\nicefrac{1}{8},$ $\nicefrac{1}{16},$ $\nicefrac{1}{32}\}$, and the corresponding widths are set to $F, 2F, 4F, 8F$, where we set $F$ to be 64 for our best model.
Regarding the DCSS, we set the hyper-parameter $L$ to be 14 and the sampling ratio $r$ to be $\nicefrac{1}{4}$.
To deliver end-to-end searching, we hand-craft a \emph{stem module} that aims to extract feature-pyramids for DCSS, and we also design a simple \emph{head block} for aggregating all of the feature-maps from DCSS to make the final prediction ( We refer the  supplemental materials for more implementation details ).

\begin{table}
\centering
\resizebox{0.95\columnwidth}{!}
{
\begin{tabular}{l | l c c c c | }
\toprule 
~~~~~~\textbf{Method} & ~~~~~~\textbf{Backbone}&~\textbf{Validation}~~&\textbf{Coarse}&\textbf{ImageNet}& ~mIoU(\%)\\
\toprule
PSPNet \cite{zhao2017pyramid}&Dilated-ResNet-101&   \ding{55}& \ding{55}&\ding{51}&78.4 \\
PSANet \cite{zhao2018psanet}     &   Dilated-ResNet-101 & \ding{51} & \ding{55}  & \ding{51}   & 80.1 \\
PADNet \cite{xu2018pad}&   Dilated-ResNet-101 & \ding{51} & \ding{55} &     \ding{51}   & 80.3 \\
Auto-DeepLab \cite{liu2019auto}&   - & \ding{51} & \ding{55} & \ding{55}   & 80.4 \\
DenseASPP \cite{yang2018denseaspp} &   WDenseNet-161 & \ding{51} & \ding{55}  & \ding{51}   & 80.6 \\
SVCNet \cite{ding2019semantic} &   ResNet-101 & \ding{51} & \ding{55} &  \ding{51}   & 81.0 \\
PSPNet \cite{zhao2017pyramid}     &     Dilated-ResNet-101 &   \ding{51} & \ding{51} & \ding{51}   & 81.2 \\
CPNet101 \cite{yu2020context} & ResNet-101 &\yes & \no & \yes & 81.3 \\
DeepLabv3 \cite{chen2017rethinking}&   ResNet-101 & \ding{51}       & \ding{51} & \ding{51}   & 81.3 \\
CCNet \cite{huang2019ccnet}     &   ResNet-101 & \ding{51}       & \ding{55}         &     \ding{51}   & 81.4 \\
DANet \cite{fu2019dual} &   Dilated-ResNet-101 & \ding{51} & \ding{55} & \ding{51}   &       81.5 \\
SpyGR \cite{li2020spatial} & ResNet-101 &\yes &\yes&\yes& 81.6\\
RPCNet \cite{zhen2020joint} & ResNet-101 &\yes &\no & \yes & 81.8 \\
\hdashline

Mapillary$^\dag$ \cite{rota2018place} &   ResNeXt-101 & \ding{51} & \ding{51}  & \ding{51} & 82.0 \\
HANet$^\dag$ \cite{choi2020cars}& ResNet-101 & \yes &\no&\yes&83.2\\
DeepLabv3+ \cite{chen2018encoder}&   Dilated-ResNet-101 & \ding{51} & \ding{51} & \ding{51}   & 82.1 \\
Auto-DeepLab \cite{liu2019auto}&   - & \ding{51} & \ding{51}  & \ding{55}   & 82.1 \\

HRNetV2 + OCR$^\dag$ \cite{wang2020deep} & HRNetV2 & \yes & \no  &\yes & 83.9 \\
DPC \cite{chen2018searching}&   Dilated-Xception-71 & \ding{51} & \ding{55}  & \ding{51} & 82.7 \\
DRN \cite{zhuang2018dense} & ResNet-101 & \ding{51} & \ding{51}  &     \ding{51}   & 82.8 \\
GSCNN$^\dag$ \cite{takikawa2019gated} &     Wide-ResNet & \ding{51}       & \ding{55} &  \ding{51}   & 82.8 \\
DecoupleSegNets$^\dag$ \cite{li2020improving}& Wide-ResNet &\yes &\no &\yes & 83.7 \\
\hdashline

DCNAS$^\dag$  &     - & \ding{55}       & \ding{55}        &     \ding{55}   &       82.8 \\
DCNAS$^\dag$  &   - & \ding{51}       & \ding{55}        &     \ding{55}   &       83.1 \\
DCNAS$^\dag$  &     - & \ding{51}       & \ding{51}        &     \ding{55}   &  83.6\\
DCNAS + ASPP$^\dag$&-& \ding{51}& \ding{51}&    \ding{55}           &       \textbf{84.3} \\
\bottomrule
\end{tabular}
}
\vspace{3pt}
\caption{\textbf{Performance on Cityscapes}. 
The table summarizes the performance on Cityscapes testing dataset.
\textbf{Validation:} Models trained with both train-fine and val-fine parts.
\textbf{ImageNet:} The backbones of model trained on ImageNet.
\textbf{Coarse:} Models that exploit extra datas in Cityscapes with coarse annotation.
\bm{$^\dag$}: Adopts the well labeled Mapillary Vistas \cite{neuhold2017mapillary} dataset.
}
\label{tb:comp_citys_test}
\end{table}

\begin{table*}[ht]
\centering
{
\resizebox{0.95\textwidth}{!}
{%
\begin{tabular}{ l | c c  c c c c c c c c c c c c c c c c c c |c}
\toprule
~~~~~Method & ~aero~ & ~bike~ & ~bird~ & ~boat~ & bottle & \,~bus~\, & \,~car~\, & \,~cat~\, & \,chair\, & \,~cow~\, & \,table\, & \,~dog~\, & \,horse\, & \,mbike\, &person & \,plant\, & \.sheep\, & ~sofa~ & \,train\, & tv & mIoU(\%)\\
\toprule
AAF \cite{ke2018adaptive}&
91.2 & 72.9 & 90.7 & 68.2 & 77.7 & 95.6 & 90.7 & 94.7 & 40.9 & 89.5 & 72.6 & 91.6 & 94.1 & 88.3 & 88.8 & 67.3 & 92.9 & 62.6 & 85.2 & 74.0 & 82.2 \\
ResNet38 \cite{wu2019wider}&
94.4 & 72.9 & 94.9 & 68.8 & 78.4 & 90.6 & 90.0 & 92.1 & 40.1 & 90.4 & 71.7 & 89.9 & 93.7 & 91.0 & 89.1 & 71.3 & 90.7 & 61.3 & 87.7 & 78.1 & 82.5 \\
DANet \cite{fu2019dual} &
        -&       -&       -&       -&       -&       -&       -&       -&       -&       -&       -&       -&       -&       -&       -&       -&       -&       -&       -&       -&    82.6 \\

PSPNet \cite{zhao2017pyramid}&
91.8 & 71.9 & 94.7 & 71.2 & 75.8 & 95.2 & 89.9 & 95.9 & 39.3 & 90.7 & 71.7 & 90.5 & 94.5 & 88.8 & 89.6 & 72.8 & 89.6 & 64.0 & 85.1 & 76.3 & 82.6 \\
DFN \cite{yu2018learning} &
        -&       -&       -&       -&       -&       -&       -&       -&       -&       -&       -&       -&       -&       -&       -&       -&       -&       -&       -&       -&    82.7 \\
EncNet \cite{zhang2018context}&
94.1 & 69.2 & 96.3 & 76.7 & \textbf{86.2} & 96.3 & 90.7 & 94.2 & 38.8 & 90.7 & 73.3 & 90.0 & 92.5 & 88.8 & 87.9 & 68.7 & 92.6 & 59.0 & 86.4 & 73.4 & 82.9 \\
PAN \cite{li2018pyramid} &
95.7 & 75.2 & 94.0 & 73.8 & 79.6 & 96.5 & 93.7 & 94.1 & 40.5 & 93.3 & 72.4 & 89.1 & 94.1 & 91.6 & 89.5 & 73.6 & 93.2 & 62.8 & 87.3 & 78.6 & 84.0 \\
CFNet \cite{zhang2019co}&
95.7 & 71.9 & 95.0 & 76.3 & 82.8 & 94.8 & 90.0 & 95.9 & 37.1 & 92.6 & 73.0 & 93.4 & 94.6 & 89.6 & 88.4 & 74.9 & \textbf{95.2} & 63.2 & \textbf{89.7} & 78.2 & 84.2 \\
APCNet \cite{he2019adaptive}&
95.8 & 75.8 & 84.5 & 76.0 & 80.6 & 96.9 & 90.0 & 96.0 & 42.0 & 93.7 & 75.4 & 91.6 & 95.0 & 90.5 & 89.3 & 75.8 & 92.8 & 61.9 & 88.9 & \textbf{79.6} & 84.2 \\
DMNet \cite{he2019dynamic} &
96.1 & \textbf{77.3} & 94.1 & 72.8 & 78.1 & \textbf{97.1} & 92.7 & 96.4 & 39.8 & 91.4 & 75.5 & 92.7 & \textbf{95.8} & 91.0 & \textbf{90.3} & \textbf{76.6} & 94.1 & 62.1 & 85.5 & 77.6 & 84.3 \\

SANet \cite{zhong2020squeeze} &95.1& 65.9 &95.4 &72.0 &80.5& 93.5 &86.8& 94.5& 40.5& 93.3& 74.6& 94.1 &- &- &- &- &-&-&-&-& 83.2\\ 

SpyGR \cite{li2020spatial} &- &-&-&-&-&-&-&-&-&-&-&-&-&-&-&-&-&- &-&-&84.2\\ 

CaC-Net\cite{liu2020learning} & 96.3& 76.2 &95.3& 78.1 &80.8& 96.5 &91.8 &96.9 &40.7 &96.3 &76.4 &94.3 &95.8 &91.3 &89.1 &73.1 &93.3 &62.2 &86.7 &80.2 &85.1 \\

\hdashline
DCNAS (Ours)&
\textbf{96.5} & 75.2 & \textbf{96.1} & \textbf{80.7} & 85.2 & 97.0 & \textbf{93.8} & \textbf{96.6} & \textbf{49.5} & \textbf{94.0} & \textbf{77.6} & \textbf{95.1} & 95.7 & \textbf{93.9} & 89.7 & 76.1 & 94.7 & \textbf{70.9} & \textbf{89.7} & 79.4 & \textbf{86.9} \\
\bottomrule
\end{tabular}
}
}
\vspace{3pt}
\caption{\textbf{Performance on PASCAL VOC 2012.} The table presents per-class semantic segmentation results on the PASCAL VOC 2012 test dataset. Our method advances the new state-of-the-art with mIoU 86.9\%.}
\label{tb:pascal_voc_no_coco}
\end{table*}

\subsection{Correlation of Performance}

In this section, we quantify the association of the model's performance during searching and training configurations.
In practice, we take the Pearson Correlation Coefficient ($\rho$) and Kendall Rank Correlation ($\tau$) Coefficient as the evaluation metric and compare our approach with contemporary works \cite{nekrasov2019fast,wu2019sparsemask,chen2019fasterseg,chen2018searching,liu2019auto}.


To make a fair comparison, we conduct the experiment about 40 times for each method. For each experiment, we run the exploration algorithm on the training dataset of Cityscapes for 80 epochs.
As the searching procedure convergences, we train the derived model for 120 epochs on the training dataset.
We report the accuracy of both stages on the validation part in each experiment for quantitative comparisons.
We shall point out that we reproduce \cite{nekrasov2019fast,wu2019sparsemask,chen2019fasterseg,liu2019auto} based on the officially released code. As to \cite{chen2018searching}, we refer the announced results from the published paper rather than reproduce the method from scratch due to the intensive computation resource requirements.


As shown in Figure \ref{fig:coef}, one can observe that our method obtains a higher validation accuracy in the searching phase compared with \cite{nekrasov2019fast,wu2019sparsemask,chen2019fasterseg,chen2018searching,liu2019auto}. This is reasonable because our method does not depend on a proxy dataset, which alleviates the gap between searching and validation data, hence leading to a better performance.
Regarding the performance when fine-tuning the derived model, our approach still achieves the best result. This is consistent with our expectation because our search space (DCSS) is more general than \cite{liu2019auto,wu2019sparsemask,chen2019fasterseg} and our gradient-based framework is easier to train compared to \cite{chen2018searching,nekrasov2019fast} that rely on reinforcement learning.
In terms of the correlation coefficient, our approach outperforms other contemporary methods.
It is not astonishing because  \cite{nekrasov2019fast,wu2019sparsemask,chen2019fasterseg,chen2018searching,liu2019auto} perform the searching with proxy, yet architectures optimized with proxy are not guaranteed to be optimal on the target task \cite{cai2018proxylessnas}.
In our case, searching and training environments share a similar protocol. Consequently, the paired validation performances shall be approximately linear to a certain extent.






\subsection{Semantic Segmentation Results}
In this section, we evaluate our optimal model structure on Cityscapes \cite{cordts2016cityscapes}, PASCAL VOC 2012 \cite{everingham2010pascal}, ADE20K \cite{zhou2017scene}, and  PASCAL-Context \cite{mottaghi2014role} datasets. 

\textbf{Cityscapes} \cite{cordts2016cityscapes} is a large-scale and challenging dataset,
following the conventional evaluation protocol \cite{cordts2016cityscapes}, we evaluate our model on 19 semantic labels  without considering the void label. 
Table \ref{tb:comp_citys_test} presents the comparison results of our method and several state-of-the-art methods. 
Without any pre-training, our base model achieves the performance with $83.1\%$ mIoU on test dataset, which outperforms most state-of-the-art methods.
In addition, we can further improve the test mIoU to $83.6\%$ when pre-training our model with the coarse data.
Exploiting ASPP \cite{chen2017rethinking} as the prediction head, our method outperforms all state of the art methods with $84.3 \%$ mIoU, including \cite{liu2019auto,li2020improving} that also utilize ASPP.

\begin{table}
\centering
\resizebox{0.9\columnwidth}{!}
{%
\begin{tabular}{l |c c |c c}
\toprule
\multicolumn{1}{c|}{\multirow{2}{*}{~~~~Methods~~~~~~~}} & \multicolumn{2}{c|}{~~~~ADE20K~~~~} & \multicolumn{2}{c}{Pascal Context}\\\cline{2-5}
& ~~~~mIoU~~~~ & ~~Pix-Acc~~ & mIoU 59-cls & mIoU 60-cls\\
\toprule

DeepLabv2 \cite{chen2017deeplab} & - & - &-& 45.7 \\
GCPNet \cite{hung2017scene} & 38.37 & 77.76 & - & 46.5 \\
RefineNet \cite{lin2017refinenet} &40.70 & -&- & 47.3 \\
MSCI \cite{lin2018multi} & - & - & -&50.3 \\
PSANet \cite{zhao2018psanet} & 43.77 & 81.51 &- &- \\
PSPNet \cite{zhao2017pyramid}& 44.94 & 81.69 &47.8 &- \\
SAC \cite{zhang2017scale}& 44.30 &-&-&-\\
CCL \cite{ding2018context}& - &-&-&51.6\\
EncNet \cite{zhang2018context} &44.65&81.69&52.6&51.7 \\
DSSPN \cite{liang2018dynamic} & 43.68 &-&-&- \\
CFNet \cite{zhang2019co}&44.89&-&54.0&-\\
CCNet \cite{huang2019ccnet}&45.22&-&-&-\\
DeepLabv3+ \cite{chen2018encoder}&45.65&82.52&-&-\\
Auto-DeepLab \cite{liu2019auto}&43.98&81.72&-&-\\
APCNet \cite{he2019adaptive} & 45.38 & - &55.6 &54.7\\
DANet \cite{fu2019dual} & - &- & - & 52.6 \\
SVCNet \cite{ding2019semantic} & - &- &53.2 & -\\
DRN \cite{zhuang2018dense} & - & - & 49.0 & - \\
SANet \cite{zhong2020squeeze} & - & - & 55.3 & 54.4 \\
SpyGR \cite{li2020spatial} & - & - & - & 52.8\\
CPNet \cite{yu2020context} & 46.27 & 81.85 & - & 53.9 \\
HRNetV2 + OCR \cite{wang2020deep} & 45.66 & - & - & \textbf{56.2} \\ 
CaC-Net \cite{liu2020learning} & 46.12 & - & - & 55.4 \\
\hdashline 
DCNAS (Ours) & \textbf{47.12} & \textbf{84.31} & \textbf{57.1} & 55.6 \\
\bottomrule
\end{tabular}
}
\vspace{3pt}
\caption{\textbf{Performance on ADE20K and PASCAL-Context.} The table presents the semantic segmentation performance on the validation part of ADE20K and PASCAL-Context according to various evaluation metrics.
}
\label{tb:ade20_val}
\end{table}

\textbf{PASCAL VOC 2012} \cite{everingham2010pascal} is another gold-standard benchmark for object segmentation, which contains 20 foreground object classes and one background class.
Following conventional \cite{chen2017deeplab,zhao2017pyramid,zhang2018context,zhang2019co,liu2019auto}, we exploit the augmented set \cite{hariharan2011semantic} by pre-training our model on the train + val parts on the augmented set, then fine-tune the model on the original PASCAL VOC 2012 benchmark.
As a result, our model advances the new state-of-the-art result to $86.9\%$ mIoU, which largely outperforms the state-of-the-art ( absolute 2.5\% mIoU improvement ). We also report the per-class comparisons with state-of-the-art mtehods in Table \ref{tb:pascal_voc_no_coco}, one can observe that our model achieves superior performance on many categories.

\textbf{PASCAL-Context} \cite{mottaghi2014role} and \textbf{ADE20K} \cite{zhou2017scene} are substantial benchmarks toward scene parsing tasks.
Being consistent with \cite{zhang2019co,huang2019ccnet,zhu2019asymmetric,liu2019auto}, we employ not only the classical mIoU but also the pixel accuracy as the evaluation metrics for ADE20K. As to PASCAL-Context, we report the mIoUs both with (60-cls) and without (59-cls) considering the background.
As shown in Table \ref{tb:ade20_val}, we present the performances of our model on the validation parts of Pascal Context and ADE20K according to corresponding evaluation metrics. Our DCNAS outperforms other state-of-the-art methods on both Pascal Context and ADE20K datasets.

It is consistent with our expectation that our approach retained leading performance on the above mentioned extensively used benchmarks with various evaluation metrics. Specifically:
(1) the proxyless searching paradigm can bridge the gap between searching and training environments, which is conducive to the discovery of promising model structures;
(2) our DCNAS includes an abundance of possible model structures, accompanied with the proxyless searching method, so that one may easily obtain an outstanding model;
(3) the multi-scale semantic features aggregation has been proved to be instrumental for visual recognization \cite{long2015fully,chen2017deeplab,zhao2017pyramid,chen2018encoder,takikawa2019gated}, and our DCNAS inherently and repeatedly applying top-down and bottom-up multi-scale features fusion, hence results in leading performance.

\subsection{Ablation Study}
In this section, we evaluate the impact of the key designs in DCNAS.
We take the mIoU(\%) on validation part of Cityscapes both in searching (S-mIoU) and training (T-mIoU) periods,  the Pearson Correlation Coefficient ($\rho$), the Kendall Rank Correlation Coefficient ($\tau$), and the searching cost (GPU Days) as the evaluation metrics.

\begin{table}
\centering
\resizebox{0.9\columnwidth}{!}
{%
\begin{tabular}{l | c c  c c }
\toprule
\textbf{Search Scheme} & $\bm{L=8}$ & $\bm{L=10}$ & $\bm{L=12}$ & $\bm{L=14}$\\
\toprule
ProxlessNAS \cite{cai2018proxylessnas} & 64.3 & 68.1 & 73.7 & 77.5\\
PC-DARTs \cite{xu2019pc}  & 70.8 & 72.4 & 76.2 & 80.6 \\
P-DARTs \cite{chen2019progressive}  & 68.5 & 70.4 & 76.9 & 79.4 \\
SPOS \cite{guo2020single} & 65.2 & 68.5 & 73.2 & 77.4\\
\hdashline
DCNAS (Ours) & \textbf{72.8} & \textbf{75.4} & \textbf{78.9} & \textbf{81.2} \\
\bottomrule
\end{tabular}
}
\vspace{3pt}
\caption{\textbf{Comparisons with State of the Art Search Methods.} The table presents the comparison results (by T-mIoU on validation set of Cityscapes) of our proxyless search paradigm with contemporary approaches \cite{cai2018proxylessnas,xu2019pc,chen2019progressive,guo2020single} under various configurations. }
\label{tb:proxylessnas}
\end{table}

Since the proxyless searching is instrumental in DCNAS, we compare our searching strategy with searching methods \cite{cai2018proxylessnas,xu2019pc,chen2019progressive,guo2020single}.  
Table \ref{tb:proxylessnas} reports the comparing results under several configurations, as a result, our proxyless searching method is more suitable for semantic image segmentation.
Besides, our searching approach obtains significant performance improvement comparing with PC-DARTs \cite{xu2019pc}.


Besides, to deeply study the impact of the proxyless search paradigm and the densely connected search space, we transfer the searching space and searching method to the Auto-DeepLab \cite{liu2019auto} framework.
As Table \ref{fig:abs_all} presents the ablation results, one may observe that, (1) the densely connected search space remains indispensable for semantic image segmentation;
(2) proxyless searching scheme is capable of bridging the gap between searching and training situations, which can further improve the performance.


\begin{table}
\centering
\resizebox{0.97\columnwidth}{!}
{%
\begin{tabular}{ c c c  c c c c }
\toprule
\textbf{Methods} & \textbf{Proxyless} & \textbf{Dense Space} & $\bm{\rho}$ & $\bm{\tau}$ & \textbf{S-mIoU} & \textbf{T-mIoU} \\
\toprule
Auto-DeepLab  & \no & \no & 0.31 & 0.21 & 34.9 & 80.3 \\
Auto-DeepLab  & \yes & \no & 0.61 & 0.45 & 37.6 & 80.6 \\
Auto-DeepLab  & \no & \yes & 0.27 & 0.19 & 49.7 & 80.7 \\
Auto-DeepLab  & \yes & \yes & 0.63 & 0.43 & 60.9 & 80.9 \\
\hdashline
Auto-DeepLab $\dag$  & \no & \no & 0.42 & 0.33 & 40.8 & 56.9 \\
Auto-DeepLab $\dag$ & \yes & \no & 0.64 & 0.48 & 43.1 & 67.5 \\
Auto-DeepLab $\dag$ & \no & \yes & 0.48 & 0.41 & 46.2 & 70.2 \\
Auto-DeepLab $\dag$ & \yes & \yes & 0.69 & 0.54 & 50.5 & 72.4 \\
\hdashline
DCNAS (Ours) & \no & \no & 0.37 & 0.25 & 45.2 & 60.1 \\
DCNAS (Ours) & \yes & \no & 0.69 & 0.56 & 51.7 & 73.3 \\
DCNAS (Ours) & \no & \yes & 0.39 & 0.32 & 61.5 & 78.9 \\
DCNAS (Ours) & \yes & \yes & 0.73 & 0.55 & 69.9 & 81.2 \\
\bottomrule
\end{tabular}
}
\vspace{3pt}
\caption{\textbf{Ablation of the Search Strategy and Search Space.} The table presents the effects of the proxyless search paradigm and the densely connected search space in both DCNAS and Auto-DeepLab frameworks, mIoUs are evaluated on validation set. 
$\bm\dag$ replace the ASPP with a simpler head used in DCNAS.
For DCNAS, 
\textbf{ \xcancel{Proxyless}} denotes that setting the sampling ratio $r=1$ in the mixture layer and uses a much lower resolution image for searching,
while \textbf{ \xcancel{Dense Space}} indicates constraining connections to the previous layers only. 
Regrading Auto-DeepLab, \textbf{Proxyless} transfer the mixture-layer and exploit higher resolution image for searching,  \textbf{Dense Space} transfer the fusion module and optimize whole candidate connections rather than the adjacent cells only.
}
\label{fig:abs_all}
\end{table}

%

\section{Discussion}
We proposed a novel NAS framework, dubbed DCNAS, to directly and proxylessly search the optimal multi-scale network architectures for dense image prediction tasks. We introduce a densely connected search space (DCSS), which contains most of the widespread human-designed network structures. With DCSS, we attempt to fully automatically search the optimal multi-scale representations of high-resolution imagery and to fuse those multi-scale features. Meanwhile, to enable the efficient searching process, we propose a novel fusion module to deal with the high GPU memory consumption and expensive computation overhead issues. 
Consequently, we implement a novel proxyless NAS framework for dense image prediction tasks.
Experiment results demonstrate that the architectures obtained from our DCNAS can surpass not only the human-invented architectures but also the automatically designed architectures from previous NAS methods the focusing on semantic segmentation task. For future work, one promising direction is investigating the lightweight neural architecture under the DCSS for the resource-constrained devices.

\clearpage
\newpage

\small{
\bibliographystyle{ieee_fullname}
\bibliography{dcnas}
}

\end{document}


\newcommand{\todo}[1]{\textcolor{red}{TODO: #1}}
\newcommand{\TODO}[1]{\textcolor{red}{\textbf{****** #1 ******}}}
\newcommand{\hongmin}[1]{{\color{orange}\textsc{Hongmin:} #1}}
\newcommand{\jianchao}[1]{{\color{magenta}\textsc{Jianchao:} #1}}
\newcommand{\rc}{\color{purple}}
\newcommand{\yes}{\ding{51}}
\newcommand{\no}{\ding{55}}
\def\etal{\emph{et al.}}

\title{DCNAS: Densely Connected Neural Architecture Search for Semantic Image Segmentation Supplementary Materials}

\author{Xiong Zhang$^{1}$, Hongmin Xu$^2$, Hong Mo$^3$, Jianchao Tan$^2$, Cheng Yang$^1$, Lei Wang$^1$, Wenqi Ren$^4$\\
\normalsize{$^1$Joyy Inc., $^2$ AI Platform, Kwai Inc., $^3$State Key Lab of VR, Beihang University, ~$^4$SKLOIS, IIE,CAS}\\
}
\maketitle

\maketitle

\section{Methodology}


\subsection{Shape Alignment Layer}
The shape-alignment layer is in a multi-branch parallel form, which aims at transforming the semantic features (with various shapes, including spatial resolutions and channel widths) to the target shape.
In practice, we use bilinear-upsampling or $3\times3$ convolutional layer with proper stride to adjust the spatial resolution and insert $1\times 1$ convolutional layer to perform channel alignment.

\subsection{Stem and Head Module}
In order to deliver end-to-end searching and training, we hand-craft a stem module that aims at extracting feature-pyramids for DCNAS, and we also design a simple prediction head for aggregating all the feature-maps that from DCNAS to make the final prediction.
In brief, the stem block consists of a $7 \times 7$ convolution and four $3 \times 3$ convolutions with stride 2, the $7 \times 7$ convolution has $32$ filters, while the last four $3 \times 3$ convolutions have $F,2F,4F,8F$ filters, respectively.
Regarding the structure of the prediction head, we concatenate the (upsampled) semantic features from $\{O_{(\nicefrac{1}{4},L)},O_{(\nicefrac{1}{8},L)},O_{(\nicefrac{1}{16},L)},O_{(\nicefrac{1}{32},L)}\}$ then apply an extra $3 \times 3$ convolution and a $1 \times 1$ convolution to make the final prediction.

\subsection{Decoding  Algorithm} 
Once the searching procedure terminates, one may derive the suitable operator for each mixture layer and the optimal architecture based on the architecture parameters $\alpha$ and $\beta$. 
For mixture layer $\ell_{(s,l)}$, we select the candidate operation that has maximum operation weight, \emph{i.e.,} 
$\mathop{\arg\max}_{o\,\in \,\mathcal{O}} \alpha^o_{(s,l)}$.
Regrading the network architecture, we utilize a  Breadth-First Search algorithm to derive the network architecture in a back to front order, and the algorithm is listed as bellow:
\begin{center}
{
\begin{minipage}{0.56\textwidth}
\centering
    \begin{algorithm}[H]
     \KwData{Architecture parameters $\beta$}
     \KwResult{Derived optimal connections $\rm opt\_conns$}
     $\rm opt\_conns:=\{\}$\;
     $\rm int\_nodes:=\{(\nicefrac{1}{4}, L),(\nicefrac{1}{8}, L),(\nicefrac{1}{16}, L),(\nicefrac{1}{32}, L)\}$\;
     \While{$\rm int\_nodes$ not empty}{
      let $\rm (s,l)=int\_nodes.pop()$\;
      \ForEach{candidate connection $\rm (s^\prime,l^\prime)\rightarrow(s,l)$} {
      	\If{ $\rm \beta_{(s^\prime,l^\prime)\rightarrow(s,l)} ) \ge 0 $}{
    		\If{$\rm l^\prime \, \textgreater \, 0$}{
    			$\rm int\_nodes.append((s^\prime,l^\prime))$\;
    		}
    		\If{$\rm (s^\prime,l^\prime)\rightarrow(s,l) \notin opt\_conns$}{
    		$\rm opt\_conns.append((s^\prime,l^\prime)\rightarrow(s,l))$
    		}
      	}
      }
     }
     \Return $\rm opt\_conns$
     \caption{Decoding Network Structure}
     \label{alg:decode_arch}
    \end{algorithm}
\end{minipage}
}
\end{center}

\subsection{Regularization Terms} 
In practice, to obtain a faster convergence, we introduce several instrumental regularization terms as bellow,
\begin{enumerate}
	\item Since we may select the optimal operator $\mathop{\arg\max}_{o\,\in \,\mathcal{O}} \alpha^o_{(s,l)}$, we introduce 
the entropy regularization term
\begin{align}
\mathcal{L}_{\alpha}=
-\,\sum_{s \in {\mathbb{S}}, l \le L} \underbrace{\sum_{o \in \mathcal{O} } w_{(s,l)}^o\ln w_{(s,l)}^o}_{\text{to derive a one-hot distribution over } \mathcal{O}}
\end{align}
 over architecture parameters $\alpha$ to derive a one-hot distribution of the operators in each mixture layer $\ell_{(s,l)}$.
	\item  The insignificant transmissions always affect the aggregating operation, which will affect the convergence.
	Thus we introduce a regularization term as: 
 \begin{align}
 \begin{split}
 \mathcal{L}_\beta &= 
\sum_{s^\prime\in \mathbb{S}, s\in \mathbb{S}, l \leq L,  l^\prime  \textless l}\, -\underbrace{\frac{1}{(1+\exp^{-\beta_{(s^\prime,l^\prime)\rightarrow(s,l)}})}}_{\text{importance of connection} {(s^\prime,l^\prime)\rightarrow(s,l)}} \cdot \quad \ln \frac{1}{(1+\exp^{-\beta_{(s^\prime,l^\prime)\rightarrow(s,l)}})}\\
&=\sum_{s^\prime\in \mathbb{S}, s\in \mathbb{S}, l \leq L,  l^\prime  \textless l}\,\ln (1+\exp^{-\beta_{(s^\prime,l^\prime)\rightarrow(s,l)}}) / (1+\exp^{-\beta_{(s^\prime,l^\prime)\rightarrow(s,l)}})
\end{split}
\end{align} 
 to help solve this challenge.
	\item We also want to constrain the minimum and maximum number of fusion modules that connect to it to be 1 and $k$, we choose to apply the Lagrangian multiplier method to reformulate such constraints as one regularization term: 
\begin{align}
\begin{split}
\mathcal{L}_{con}&=\sum_{s\in\mathbb{S}, l \le L} 
	\max(1-\underbrace{\sum_{s^\prime\in\mathbb{S}, l^\prime  \,<\, l} \frac{1}{1+\exp^{-\beta_{(s^\prime,l^\prime)\rightarrow(s,l)}}}}_{\text{in degree}},0) \\
	&+ \, \sum_{s\in\mathbb{S}, l \le L} \max(\underbrace{\sum_{s^\prime\in\mathbb{S}, l^\prime  \,<\, l} \frac{1}{1+\exp^{-\beta_{(s^\prime,l^\prime)\rightarrow(s,l)}}}}_{\text{in degree}}-k,0).
	\end{split}
	 \end{align}
\end{enumerate}

\subsection{Searching Protocols}
We conduct architecture search on the Cityscapes dataset \cite{cordts2016cityscapes}, more specifically,  we divide the training samples with fine annotations into two parts,  with $2, 000$ and $975$ samples respectively.
 The $2, 000$ training samples are used for updating convolutional weights $w$ while the $975$ samples are used for optimizing architecture parameters $\{\alpha,\beta\}$. 
Besides we make use of the $500$ fine annotated validation samples for model selection.
For searching configurations, we set the initial learning rate as $0.01$ and $0.0005$ for $w$ and $\{\alpha,\beta\}$ respectively, 
schedule the learning rate with polynomial policy with factor $(1-(\frac{i t e r}{i t e r_{m a x}})^{0.9})$,
apply weight decay for $w$ with coefficients $0.0001$.
Following conventional data augmentation paradigm, we resize the image with random scale $[0.5, \,2.0]$, then random crop a patch with size $1024 \times 512$, apart from that, we do not employ any augmentation tricks though which shall further improve the performance.
Consider that the architecture parameters are hard to optimize,  we exploit the Adam optimizer to update $\{\alpha, \beta\}$, while for convolutional weights $w$, we adopt the general SGD for sake of better convergence.
The searching procedure takes 1.4 days for 120 epochs on 4 GPUs.

\subsection{Training Protocols}
We evaluate our optimal model on Cityscapes \cite{cordts2016cityscapes}, PASCAL VOC 2012 \cite{everingham2010pascal}, ADE20K \cite{zhou2017scene}, and  PASCAL-Context \cite{mottaghi2014role} datasets. 
We share the same training protocols for all the datasets mentioned above, except for particular dataset-specific configurations.
Specifically, during training we set the initial learning rate to $0.01$ and adopt the polynomial scheduler \cite{liu2015parsenet} to drop the learning rate, we exploit the syncBN \cite{peng2018megdet,liu2018path} to synchronize the mean and variance across different GPUs,  we conduct the experiments with batch size 32 on 4 GPUs, and we use the SGD with momentum 0.7 to optimize the convolutional weights.
Data are augmented by random scaling in range of [0.5, 2.0], random horizontal flipping, and random cropping, where we set the crop size $1024 \times 512 $ for Cityscapes and $513 \times 513$ for others.
Considering the complexity and scale of the benchmarks, empirically, we train the model on tough Cityscape for 800 epochs, concerning other benchmarks, we set the training epochs to 240. 

\subsection{Correlation of Performance}
We utilize the  Pearson Correlation Coefficient ($\rho$)  and Kendall Rank Correlation ($\tau$) Coefficient to quantify the correlation of performance between searching and training situations.
Denote $(X, Y)$ is a two dimension random variables, and $\{x_i, y_i\}_{i=1}^{n}$ are $n$ observations, the we may estimate the Pearson Correlation Coefficient ($\rho$) as,
\begin{align}
\begin{split}
\rho&=\frac{COV(X,Y)}{\sqrt{D(X)  D(Y)}} \\
&=\frac{E(XY)-E(X)E(Y)}{\sqrt{D(X)D(Y)}} \\
\end{split}
\end{align}
and we may calculate the Kendall Rank Correlation ($\tau$) Coefficient with, 
\begin{align}
\tau=\frac{\sum_{i \le n, j \textless i} \mathds{1}\{x_i \textless x_j\}  \mathds{1}\{y_i \textless y_j \} - \sum_{i \le n, j \textless i} \mathds{1}\{x_i \textless x_j\}  \mathds{1}\{y_i \textgreater y_j \}}{
\left (\begin{array}{c}
n \\
2
\end{array}
\right )
}
\end{align}

\begin{table}
\centering
\resizebox{0.4\columnwidth}{!}
{
\begin{tabular}{cccc}
\toprule
\textbf{Methods} & \textbf{\#Params} & \textbf{FLOPs} & \textbf{FPS} \\
\toprule
Auto-DeepLab & 44.42M & 695.03G &- \\
\hdashline
DCNAS (Ours) & 21.49M & 294.57G & 6.3 \\
\bottomrule
\end{tabular}
}
\vspace{3pt}
\caption{\textbf{Comparison with Auto-DeepLab.} FPS is evaluated on 1080Ti GPU with CUDA8.0 (inputs with 2048 x 1024).
}
\label{tb:compare}
\end{table}

\section{Experiments}
\subsection{Model Efficiency}
Since we do not guide the searching process to derive a sparse and compact model, therefore, the derived model tends to uses all nodes, and searching for sparse while compact structures for realtime semantic image segmentation is indeed our further work.
We also compare our optimal model with contemporary work of \cite{liu2019auto, chen2018searching}, as Table \ref{tb:compare} presents the comparison results,  our method achieves better performance with lower FLOPs.  

\subsection{Qualitative Results}
To better understand the capability of DCNAS, we drawn several examples from the validation set (Figure \ref{fig:citys_val}) and the testing part (Figure \ref{fig:citys_test}). 
The DCNAS can produce precise predictions about small targets  (\emph{e.g.,} tsign, pole)  in the scene thanks to the ability to capture subtle details in high-resolution imageries.
Meanwhile, owing to the capacity to extract long-range global information, DCNAS is good at estimating the segmentation mask of massive while complicated objects  (\emph{e.g.,} bus, truck, terrain, sidewalk, rider, building).

\subsection{Ablation Study}
To better understand the impact of different design choices, we evaluate our framework in various settings.
Two main design choices exist in this work, the impact of hyper-parameters $L$ and $k$, and the effect of the novel regularization terms.
We still take the mIoU as the evaluation metric for searching (S-mIoU) and training (T-mIoU) periods on the validation dataset of the Cityscapes.

\textbf{Operator Space $\mathcal{O}$.} In our work, we use a totally different operator space compared with Auto-DeepLab \cite{liu2019auto}. To make an apple to apple comparison, we evaluate the performance of DCNAS with the same operator space in \cite{liu2019auto}, and Table \ref{abs:mbv3} reveals that MobileNetV3 operator space do not substantially affect both the performances and the correlation coefficients.

\begin{table}
\centering
        \begin{tabular}{cccccc}
        \toprule
	\textbf{Methods}&\textbf{MobileNetV3}&~~~\bm{$\rho$}~~~&~~~$\bm\tau$ ~~~&\textbf{S-mIoU} &\textbf{T-mIoU}\\
        \toprule
        	Auto-DeepLab & \no & 0.31 & 0.21 & 35.1 & 80.3 \\
	Auto-DeepLab & \yes & 0.34 & 0.24 & 35.0 & 80.4 \\
	\hdashline
	DCNAS (Ours) & \no & 0.74  & 0.53 &69.7 & 81.0 \\
	DCNAS (Ours) & \yes & 0.73 & 0.55 &69.9 & 81.2\\
	\bottomrule
        \end{tabular}
    \vspace{3pt}
	\caption{\textbf{Ablation Study.} The table investigate the impact ofs MobileNetV3 operator space.}
		            \label{abs:mbv3}

\end{table}
\begin{table}
\centering
\begin{tabular}{cccc}
\toprule
\textbf{Budget} & \textbf{Regularizers} & \textbf{S-mIoU} & \textbf{T-mIoU} \\
\toprule
$1\times $ & \yes & 69.9 & 81.2 \\
$1\times $ & \no & 48.1 & 76.5 \\
$2\times $ & \no & 60.2 & 78.9 \\
$3\times $ & \no & 67.5 & 80.7 \\
$4\times $ & \no & 70.0 & 81.4\\
\bottomrule
\end{tabular}
\caption{\textbf{Ablation Study.} We explore the impact of regularizers over performance under several searching budgets (GPU Days).}
\label{tb:abs_reg}
\end{table}

\begin{table}
\centering
\resizebox{0.4\columnwidth}{!}
{
\begin{tabular}{ccccc}
\toprule
~$\bm{\mathcal{L}_{con}}$~~&~~$\bm{\mathcal{L}_\beta}$~~&~~$\bm{\mathcal{L}_\alpha}$~~&~~\textbf{S-mIoU}~ &~\textbf{T-mIoU}~~\\
\toprule
	\no & \no & \no  & 48.1  &76.5\\
	\no & \no & \yes& 51.3  & 77.1\\
	\no & \yes& \no& 63.9   & 79.4\\
	\no & \yes& \yes & 65.3 & 79.9\\
	\yes & \no & \no & 55.2  & 78.3\\
	\yes & \no & \yes & 58.7 & 78.4\\
	\yes  & \yes & \no & 69.6 & 80.9\\
	\yes& \yes& \yes& 69.9   & 81.2\\
\bottomrule
\end{tabular}
}
    \vspace{3pt}
	\caption{\textbf{Ablation Study.} We investigate the impact of different regularization terms by comparing the performance on Cityscapes validation set, in which  $L$ and $k$ are set to be 14 and 3.}
		            \label{abs:reg}

\end{table}

\begin{table}
\centering
\resizebox{0.4\columnwidth}{!}
{%
\begin{tabular}{c c c  c c }
\toprule
$\bm{L}$ & $\bm{r}$ & \textbf{GPU Days} & \textbf{S-mIoU}& \textbf{T-mIoU}\\
\toprule
8 & $1$ & 8.6 & 61.4 & 73.1 \\
8 &  $\nicefrac{1}{2}$ & 5.2 & 60.9 & 73.0 \\
8 &  $\nicefrac{1}{4}$ & 2.8 & 60.1 & 72.8 \\
8 &  $\nicefrac{1}{8}$ & 1.5 & 57.9 & 71.1 \\
8 &  $\nicefrac{1}{16}$ & 0.9 & 55.2 & 68.3 \\
\hdashline
14 &  $1$ & 16.2 & 71.1 & 81.31 \\
14 &  $\nicefrac{1}{2}$ & 9.8 & 70.8 & 81.3 \\
14 &  $\nicefrac{1}{4}$ & 5.6 & 69.9 & 81.2 \\
14 &  $\nicefrac{1}{8}$ & 3.0 & 67.6 & 80.6 \\
14 &  $\nicefrac{1}{16}$ & 1.9 & 63.8 & 78.9 \\
\bottomrule
\end{tabular}
}

\caption{\textbf{Ablation of Sampling Ratio.} The table presents the impact of sampling ratio $r$ in mixture layer.}
\label{tb:sampling_ratio}
\end{table}

\begin{table}
      \centering
        \begin{tabular}{cccc}
        \toprule
            ~~~$\bm{L}$~~~~&~~~~$\bm{k}$~~~~&~~~\textbf{S-mIoU}~~ & ~~\textbf{T-mIoU}~~\\
        \toprule
	14 & 1 & 59.4 &  76.2\\
        14 & 2 & 66.8 & 79.5\\
        14 & 3 & \textbf{69.7} & \textbf{81.2}\\        
        14 & 4 & 69.6 & 81.0\\
        14 & 5 & 69.3 & 80.9\\
	\bottomrule
        \end{tabular}
        
    \vspace{3pt}
        \caption{\textbf{Ablation of $k$.} We present the performance of our DCNAS on Cityscapes validation dataset with varying configurations of $k$.}
		\label{abs:lks}
\end{table}

\begin{table}
\centering
\resizebox{0.75\columnwidth}{!}
{%
\begin{tabular}{l c c c | c c c}
\toprule
\multicolumn{1}{c}{\multirow{2}{*}{~~~~Methods~~~~}} &
 \multicolumn{3}{c|}{~~higher is better} &
  \multicolumn{3}{c}{lower is better}
 \\\cline{2-7}
& ~~$\delta<1.25$~&~$\delta<1.25^2$~&~$\delta < 1.25^3$~~&~~AbsRel~~&~~RMSE~~&~~~log10~~\\
\toprule
Make3D \cite{saxena2008make3d} & 0.447 & 0.745 & 0.897 & 0.349 & 1.214 &- \\
Joint HCRF \cite{wang2015towards} & 0.605 & 0.890 & 0.970 & 0.220 & 0.824 &- \\
Liu \emph{et al.} \cite{liu2015learning} & 0.650 & 0.906 & 0.976 & 0.213 & 0.759 & 0.087 \\
Eigen \emph{et al.} \cite{eigen2015predicting} & 0.769 & 0.950 & 0.988 & 0.158 & 0.641 & - \\
Li \emph{et al.} \cite{li2017two} & 0.789 & 0.955 & 0.988 & 0.152 & 0.611 & 0.064 \\
Gur \emph{et al.} \cite{gur2019single} & 0.797 & 0.951 & 0.987 & 0.149 & 0.546 & 0.063\\
Chakrabarti \emph{et al.} \cite{chakrabarti2016depth} & 0.806 & 0.958 & 0.987 & 0.149 & 0.620 & -\\
Laina \emph{et al.} \cite{laina2016deeper} & 0.811 & 0.953 & 0.988 & 0.127 & 0.573 & 0.055 \\
Xu \emph{et al.} \cite{xu2017multi} & 0.811 & 0.954 & 0.987 & 0.121 & 0.586 & 0.052 \\
Lee \emph{at al.} \cite{lee2018single} & 0.815 & 0.963 & 0.991 & 0.139 & 0.572 & - \\

DORN \cite{fu2018deep}& 0.828 & 0.965 & 0.992 & 0.115 & 0.509 & 0.051 \\
Geonet \cite{qi2018geonet} & 0.834 & 0.960 & 0.990 & 0.128 & 0.569 & 0.057 \\
Lee \emph{et al.} \cite{lee2019monocular} & 0.837 & 0.971 & 0.994 & - & 0.538 & - \\
Yin \emph{et al.}  \cite{yin2019enforcing} & 0.875 & 0.976 & 0.994 & 0.108 & \textbf{0.416} & \textbf{0.048} \\
Freeform \cite{chang2019deep} & \textbf{0.930} & \textbf{0.990} & \textbf{0.999} & \textbf{0.087} & 0.433& 0.052 \\
\hdashline
Ours (DCNAS) & 0.836 & 0.968 & 0.994 & 0.113 & 0.497 & 0.052  \\
\bottomrule
\end{tabular}
}
\caption{\textbf{Quantitative performance on NYU Depth v2.} Our method achieves competitive result against state-of-the-art approaches. Though not state-of-the-art, we argue it is reasonable when considering the scale of NYU benchmark and the saturated performance.
}
\label{tb:nyuv2}
\end{table}

\textbf{Hyperparameter $\bm{k}$.} $k$ is an important parameter in regularization term $\mathcal{L}_{con}$, as presents in In Table \ref{abs:lks}, as $k$ increases, the performance grows first and then decreases. Because, (1) 
small $k$ will reduce the search space largely, which may miss the optimal architecture and results in poor performance,
(2) large $k$ will lead a tremendous search space, from which explore promising architecture is challenging.
As a compromise, we choose $k=3$ in our experiments.

\textbf{Regularization Terms.} Regarding the regularization terms, we observe that the $\mathcal{L}_\beta$ is crucial to performance improvement, and adding the regularization term $\mathcal{L}_{con}$ can yield marginal performance improvement. We think there are two reasons: (1) the regularization term $\mathcal{L}_\beta$ forces the relaxed continuous representations of path connection to be 0 or 1,  which stabilizes the network architecture;
(2) constraint $\mathcal{L}_{con}$ helps prune insignificant path-level connections and derive a sparse model structure, which further stabilizes the network architecture.
Table \ref{tb:abs_reg} exhibits that those regularization terms is capable of reducing the searching cost substantially.

Moreover, as presented in Table \ref{abs:reg}, adding the regularization term $\mathcal{L}_\alpha$ can further improve the performance, since $\mathcal{L}_\alpha$ leads to a discrete distribution of operator space in each mixture-layer, which yields more stable and reliable training.

\textbf{Sampling Ratio $r$ in Mixture Layer.} We also investigate the impact of the sampling ratio $r$ in mixture layer. As Table \ref{tb:sampling_ratio} demonstrates that   $\nicefrac{1}{4}$ is a good trade-off between search efficiency and model accuracy.

\subsection{Generalization Capability} To evaluate the generalization performance of our approach, we evaluate the DCNAS on task of monocular depth estimation on NYU Depth v2 \cite{Silberman:ECCV12}, as Table \ref{tb:nyuv2} and Figure \ref{fig:nyu_test} present the quantitative and qualitative results. 
It is exciting that the quantitative and qualitative experiment reveals our DCNAS can generalize well to other dense image prediction task.

\begin{figure}
\centering
\includegraphics[width=0.82\textwidth]{pdfs/citys_val.jpg}
\caption{\textbf{Qualitatively results on cityscapes validation part.} We present several prediction results on Cityscapes validation part. The three columns represent the RGB input, prediction given by DCNAS, and the ground-truth, respectively.}
\label{fig:citys_val}
\end{figure}

\begin{figure}
\centering
\includegraphics[width=0.82\textwidth]{pdfs/citys_test.jpg}
\caption{\textbf{Qualitatively results on Cityscapes test part.} We present several prediction results on Cityscapes test part. In which each row comprise two samples, each sample contains the RGB input and the result produced by DCNAS.}
\label{fig:citys_test}
\end{figure}

\begin{figure}
\centering
\includegraphics[width=0.9\textwidth]{pdfs/nyu_test.jpg}
\caption{\textbf{Qualitatively results on NYU Depth v2.} We present several prediction results on NYU test part. In which each row comprise two samples, each sample contains the RGB input, the mask result produced by DCNAS, and the ground truth, respectively.}
\label{fig:nyu_test}
\end{figure}

\clearpage
\newpage

\bibliographystyle{ieee_fullname}
\bibliography{dcnas.bib}